\documentclass{IEEEtran}
\usepackage{mathtools,amssymb,commath}
\usepackage{amsmath}
\usepackage{graphicx,import}
\usepackage{verbatim}
\usepackage{color}
\usepackage{hyperref}
\hypersetup{colorlinks=true}  
\usepackage{subcaption}
\usepackage{tikz}
\usepackage{booktabs}
\usepackage{algorithmic} 
\usepackage[linesnumbered,ruled,vlined]{algorithm2e}
\DeclareMathOperator*{\minimize}{minimize}				%

\SetCommentSty{mycommfont}
\SetKwInput{KwInput}{Input}
\SetKwInput{KwOutput}{Output}

\title{\LARGE \bf Multi-contact Stochastic Predictive Control for Legged Robots with Contact Locations Uncertainty}

\author{Ahmad Gazar$^{1,4}$, Majid Khadiv$^{2}$, Andrea Del Prete$^{3}$, Ludovic Righetti$^{4}$ 
\thanks{
This work was partially supported by the European Union's Horizon 2020 research and innovation program under Grant Agreement 780684 and the US National Science Foundation under Grant CMMI-1825993.
The authors thank the International Max Planck Research School for Intelligent Systems (IMPRS-IS)
for the non-financial support of Ahmad Gazar. }
\thanks{$^{1}$ Max Planck Institute for Intelligent Systems, Tuebingen, Germany. {\tt\small ahmad.gazar@tuebingen.mpg.de}}%
\thanks{$^{2}$ Munich Institute of Robotics and Machine Intelligence, Technical University of Munich, Germany. {\tt\small majid.khadiv@tum.de}}%
\thanks{$^{3}$ Industrial Engineering Department, University of Trento, Italy. {\tt\small andrea.delprete@unitn.it}}%
\thanks{$^{4}$ Tandon School of Engineering, New York University, New York, USA. {\tt\small ludovic.righetti@nyu.edu}}%
}
\begin{document}
    \maketitle
    \thispagestyle{empty}
    \pagestyle{empty}
    \begin{abstract}
        Trajectory optimization under uncertainties is a challenging problem for robots in contact with the environment. Such uncertainties are inevitable due to estimation errors, control imperfections, and model mismatches between planning models used for control and the real robot dynamics. This induces control policies that could violate the contact location constraints by making contact at unintended locations, leading to unsafe motion plans. This work addresses the problem of robust kino-dynamic whole-body trajectory optimization using stochastic nonlinear model predictive control (SNMPC) by considering additive uncertainties on the model dynamics subject to contact location chance constraints as a function of the robot's full kinematics. We demonstrate the benefit of using SNMPC over classic nonlinear MPC (NMPC) for whole-body trajectory optimization in terms of contact location constraint satisfaction (safety). We run extensive Monte-Carlo simulations for a quadruped robot performing agile trotting and bounding motions over small stepping stones, where contact location satisfaction becomes critical. Our results show that SNMPC can perform all motions safely with $100\%$ success rate, while NMPC failed $48.3\%$ of all motions.    
    \end{abstract}
    \vspace{-0.15 cm}
\section{Introduction}
\vspace{-0.15 cm}
\label{sec:intro}
Trajectory optimization for robots in contact-rich scenarios poses several control challenges due to the hybrid nature of their under-actuated dynamics that needs to be stabilized through constrained contact forces at desired contact locations with the environment~\cite{carpentier2021recent}~\cite{wensing2022optimization}. Model Predictive Control (MPC) has been a favorable tool of choice for trajectory optimization as it exploits the causal structure of the rolled-out dynamics while guaranteeing constraint satisfaction~\cite{rawlings2017model}. 
\\
\indent Despite the inherent robustness of MPC for disturbance rejection through high-frequency re-planning, dealing with persistent disturbances remains critical for the successful execution of agile motions for legged robots. Such disturbances might arise from estimation errors, model mismatches, or imperfect controls. This induces control policies that misplace end-effectors at unintended contact locations, leading to failed motions. Thus, planning under uncertainties becomes a necessity for safe trajectory optimization.
\\
\indent Recently, planning under uncertainties gained more attention in the legged robotics community. For example, Villa et. al~\cite{villa2017} used a tube-based Robust MPC (RMPC) by taking into account additive uncertainties on the dynamics using a Linear Inverted Pendulum Model (LIPM) for bipedal walking, and designing robust Center of Pressure (CoP) constraints that accommodate for all disturbance realizations inside of the CoP disturbance set. However, RMPC approaches are rather conservative and tend to sacrifice performance to guarantee robustness. To this end, we resorted to a less conservative chance-constrained stochastic MPC formulation by treating the additive uncertainties stochastically, and satisfying state constraints in a probabilistic sense known as \textit{chance-constraints}~\cite{gazar2020}. Although LIPM allows the application of linear SMPC approaches, it limits the range of agile motions and relevant uncertainties to be considered for legged robots, and their effect on contact location constraint satisfaction. 
\\
\indent For optimizing whole-body motions, indirect methods like iLQR/Differential Dynamic Programming (DDP)~\cite{mayne1973differential} have become a popular choice in the robotics community~\cite{tassa2012synthesis}~\cite{mastalli2020crocoddyl}. To incorporate uncertainties in DDP formulations, Morimoto et al. considered a minimax DDP  for simple bipedal walking dynamics subject to additive disturbances on the viscous friction of the robot joints~\cite{morimoto2003}. Recently~\cite{hammoud2022, jordana2022} used risk-sensitive DDP that accounts for process and measurement uncertainties. Other lines of work resorted to sampling-based methods to approximate the stochastic optimal control problem. For instance, Mordatch et al. used an ensemble of perturbed models that allowed them to transfer the control policy to a humanoid robot~\cite{mordatch2015}. However, sampling-based approaches are computationally expensive for real-time MPC applications of high-dimensional robotic systems. Despite the different risk measures adopted in the above approaches, they do not include constraints in their formulations, which is essential for safe trajectory optimization.
\\
\indent Other approaches incorporated uncertainties through contact by solving a Stochastic Linear Complementarity Problem (SLCP). For example,~\cite{tassa2010stochastic} solved a SLCP to avoid the discontinuities of the complementarity problem. This allowed them to optimize smoothly through contacts by offering a trade-off between contact complementarity accuracy, and the feasibility of the problem. Despite the success of DDP approaches, they do not consider the effect of uncertainties on constraint satisfaction, which is crucial in robotics. Finally, Drnach et al. used a direct contact-implicit approach to solve a SLCP with chance constraints~\cite{drnach2021}. Due to the nature of the mixed-integer OCP in contact-implicit approaches, they are hard to solve and are best suited for offline optimization.     
\\
\indent Some of the limitations of previous approaches are: 1) they do not consider explicitly the effect of uncertainty on constraint satisfaction, which is the case in most aforementioned DDP approaches. 2) Contact-implicit approaches are usually hard to tune and get easily stuck in local minima, which limits their applicability for MPC. 3) Unlike stochastic trajectory optimization, robust approaches are conservative as they sacrifice performance for safety. This work addresses the above limitations with the following contributions: 
\begin{itemize}
    \item We solve for the first time a stochastic kino-dynamic whole-body trajectory optimization subject to additive uncertainties on the dynamics. Contrary to our previous work on stochastic centroidal momentum trajectory optimization~\cite{gazar2023} which splits the OCP between the centroidal dynamics and whole-body dynamics, we optimize for both the centroidal momentum dynamics and the full robot kinematics in one OCP. This allows us to model uncertainties on the optimized contact locations rather than being fixed parametric contact location uncertainties. Moreover, in this work we solve the OCP in a receding horizon fashion using a real-time iteration scheme rather than an offline stochastic trajectory optimization with full convergence as in~\cite{gazar2023}.  
    \item We design contact location chance constraints inside an approximate real-time SQP-type iteration. This is less conservative than considering worst-case disturbance in robust optimization, where constraints are to be satisfied for all possible realizations. Instead, we satisfy constraints in a probabilistic sense, while maintaining the same computational complexity as NMPC without degrading the performance. 
    \item We compared SNMPC against NMPC by running Monte-Carlo simulations of the quadruped robot Solo for dynamic trotting and bounding gaits on a challenging non-coplanar terrain. Furthermore, We report the robustness induced by SNMPC against a heuristic-based NMPC (HNMPC), where the contact location constraints were shrunk by hand using a heuristic safety margin. Our results show that SNMPC performed all motions with a $100\%$ success rate, while NMPC and HNMPC failed $48.3\%$ and $47.6\%$ of the time.  
\end{itemize}

    \newtheorem{Problem}{Problem}
\newtheorem{assumption}{Assumption}
\section{Background}
\label{sec:background}
\textbf{Notation:} A random variable $x$ following a distribution $\mathcal{Q}$ is denoted as $x \sim \mathcal{Q}$, with $\mathbb{E}[x]$ being the expected value of $x$, and $\boldsymbol{\Sigma}_x \triangleq \mathbb{E}[(\boldsymbol{x}-\mathbb{E}[\boldsymbol{x}])(\boldsymbol{x}-\mathbb{E}[\boldsymbol{x}])^\top]$, and the weighted $l_2$ norm is $\norm{\boldsymbol{y}}_{\boldsymbol{P}} \triangleq \boldsymbol{y}^\top \boldsymbol{P} \boldsymbol{y}$. Symbols $\bigvee$, and $\bigwedge$ represent logical conjunction and disjunction respectively. 
\subsection{Multi-Contact Robot Dynamics}
The full-body dynamics of a floating-base robot can be derived using Euler-Lagrange equations of motion 
\begin{IEEEeqnarray}{LLL}
\boldsymbol{M}(\boldsymbol{q})\ddot{\boldsymbol{q}} + \boldsymbol{b}(\boldsymbol{q}, \dot{\boldsymbol{q}}) = \sum^{n_c}_{i=0} \boldsymbol{J}^\top_i(\boldsymbol{q})\boldsymbol{\lambda}_{i} +\boldsymbol{S}^\top \boldsymbol{\tau}.  \label{eq:manipulator dynamics}      
\end{IEEEeqnarray}
The generalized robot position $\small \boldsymbol{q} = \begin{bmatrix} \boldsymbol{x}_b^\top, \boldsymbol{\theta}_j^\top \end{bmatrix}^\top \in \mathbb{SE}(3) \times \mathbb{R}^{n_j}$ represents the robot's floating base pose, and joint positions respectively. The inertia matrix is denoted as $\boldsymbol{M}(\boldsymbol{q}) \in \mathbb{R}^{(6+n_j) \times (6+n_j)}$, and $\boldsymbol{b}(\boldsymbol{q}, \dot{\boldsymbol{q}}) \in \mathbb{R}^{6+n_j}$ is the vector capturing the Coriolis, centrifugal, gravity and joint friction forces. $\boldsymbol{J}_{i}$ represents the  associated jacobian of the $i$-th end-effector contact force $\boldsymbol{\lambda}_{i} \in \mathbb{R}^3$ acting on the environment for a point-foot robot. Finally, $\small \boldsymbol{S} = \begin{bmatrix}\boldsymbol{0}_{(n_j \times 6)},\, \boldsymbol{I}_{n_j} \end{bmatrix}$ is the selector matrix of the actuated joint torques $\boldsymbol{\tau}$. By focusing on the under-actuated part of the dynamics in \eqref{eq:manipulator dynamics} (i.e. first 6 equations), one can plan centroidal momentum trajectories by exploiting the relationship between the linear momentum $\boldsymbol{l} \in \mathbb{R}^3$ and angular momentum $\boldsymbol{\kappa} \in \mathbb{R}^3$ about the CoM, and the generalized robot velocities $\dot{\boldsymbol{q}}$ as $\dot{\boldsymbol{h}}_\mathcal{G} =  \small \begin{bmatrix}\dot{\boldsymbol{\kappa}}, \dot{\boldsymbol{l}}\end{bmatrix}^\top = \boldsymbol{A}_\mathcal{G}(\boldsymbol{q}) \ddot{\boldsymbol{q}} + \dot{\boldsymbol{A}}_\mathcal{G} (\boldsymbol{q}) \dot{\boldsymbol{q}}$
via the \textit{Centroidal Momentum Matrix} (CMM) \mbox{$\boldsymbol{A}_\mathcal{G} \in \mathbb{R}^{6 \times (6+n_j)}$}~\cite{orin2008}. With the same spirit as~\cite{dai2014whole-body}, we are interested in planning kino-dynamic whole-body motions using centroidal momentum dynamics and full robot kinematics (2nd order kinematics) as follows:  
\begin{IEEEeqnarray}{LLL}
\label{eq:path integral constraints}
    \underbrace{
        \frac{d}{d t}
        \begin{bmatrix} 
            \boldsymbol{c}\\ 
            \boldsymbol{l}\\  
            \boldsymbol{\kappa}\\          
        \end{bmatrix} 
        = 
        \begin{bmatrix}
            \frac{1}{m}\boldsymbol{l} \\ 
            m \boldsymbol{g} + \sum^{n_c}_{i=1} \boldsymbol{\lambda}_i \\ 
            \sum^{n_c}_{i=1} \left(\textbf{FK}_i(\tilde{\boldsymbol{q}}\right) - \boldsymbol{c}) \times \boldsymbol{\lambda}_i   
        \end{bmatrix},
    }_{\text{Centroidal momentum dynamics}} 
    \IEEEyesnumber\IEEEyessubnumber \label{eq:centroidal momentum dynamics}
    \\
    \nonumber 
    \\
    \underbrace{
        \frac{d}{d t}
        \begin{bmatrix} 
            \boldsymbol{p}_b\\ 
            \Delta \boldsymbol{q}_b\\  
            \boldsymbol{\theta}_j\\ 
            \boldsymbol{v}_b\\ 
            \boldsymbol{\omega}_b\\  
            \boldsymbol{v}_j\\ 
        \end{bmatrix} 
        =
        \begin{bmatrix} 
            \boldsymbol{v}_b\\ 
            \boldsymbol{\omega}_b\\  
            \boldsymbol{v}_j\\ 
            \boldsymbol{a}_b\\ 
            \boldsymbol{\psi}_b\\  
            \boldsymbol{a}_j\\ 
        \end{bmatrix},
        \quad
        \tilde{\boldsymbol{q}} \triangleq
        \begin{bmatrix}
        \boldsymbol{p}_b \\
        \frac{1}{2}\boldsymbol{q_{\text{ref}_b}} \otimes \Delta \boldsymbol{q}_b \\
        \boldsymbol{\theta}_j
        \end{bmatrix}
    }_{\text{Full robot kinematics}}. \IEEEyessubnumber \label{eq:double integrator kinematics}
\end{IEEEeqnarray}
$\boldsymbol{c} \in \mathbb{R}^3$ represents the CoM of the robot, with $m$ being the total mass of the robot subject to the gravity vector $\boldsymbol{g}$. The forward kinematics function $\textbf{FK}_i(.): \mathbb{Q} \mapsto \mathbb{R}^3$ computes the $i$-th end-effector's contact position for a given robot configuration. For the simplicity of dynamics integration and constraints linearization later, we choose to optimize for the relative base orientation $\Delta \boldsymbol{q}_b$ w.r.t. an absolute base reference $\boldsymbol{q}_{\text{ref}_b}$ instead of $\boldsymbol{q}_b$ directly. $\boldsymbol{p}_b$ and $\boldsymbol{v}_b \in \mathbb{R}^3$ are the base linear position and velocity, while $\boldsymbol{\omega}_b$ and $\boldsymbol{\psi}_b \in \mathbb{R}^3$ are the base angular velocity and acceleration respectively. Finally, we \textit{transcribe} the above continuous dynamics using direct collocation into the following MPC problem with pre-specified contact mode and timing $\Delta_k$. The state and control optimizers at the $k$-th discretization step are 
$\small \boldsymbol{x}_k = 
    \begin{bmatrix}
    \boldsymbol{c}_k^\top,\, \boldsymbol{l}_k^\top, \,\boldsymbol{\kappa}_k^\top,\, \boldsymbol{p}^\top_{b_k},\, \Delta\boldsymbol{q}^\top_{b_k},\, \boldsymbol{\theta}^\top_{j_k}, \boldsymbol{v}^\top_{b_k},\, \boldsymbol{\omega}^\top_{b_k},\, \boldsymbol{v}^\top_{j_k}
    \end{bmatrix}^\top \in \mathbb{R}^n$, and $\small \boldsymbol{u}_k  = 
    \begin{bmatrix}
    \boldsymbol{\lambda}_{i,k}^\top,\, \dots, \,\boldsymbol{\lambda}_{n_c, k}^\top,\, \boldsymbol{a}^\top_{b_k},\, \boldsymbol{\psi}^\top_{b_k},\, \boldsymbol{a}^\top_{j_k}
    \end{bmatrix}^\top \in \mathbb{R}^m$ 
with $ n = 21 + 6n_c$, and $m = 6 + 6n_c$ for point feet robots.
\begin{Problem} Kino-Dynamic NMPC Problem
    \label{NMPC problem}
    \begin{IEEEeqnarray}{lcr}
        \minimize_{\substack{\boldsymbol{X}, \boldsymbol{U}, \boldsymbol{S}}} \, \mathcal{L}_{\mathrm{total}}(\boldsymbol{X}, \boldsymbol{U}, \boldsymbol{S})  \label{eq:nominal cost function} \IEEEyesnumber\IEEEyessubnumber
        \\
      \,\, \mathrm{s.t.} \nonumber
      \\
      \quad \boldsymbol{f}_{\mathrm{impl}}(\boldsymbol{x}_k, \boldsymbol{x}_{k+1}, \boldsymbol{u}_k) = \boldsymbol{0}, 
        \IEEEyessubnumber \label{eq:discrete dynamics} 
        \\
        \quad \boldsymbol{h}(\boldsymbol{x}_k, \boldsymbol{u}_k) + \boldsymbol{J}_{\mathrm{sh}} \boldsymbol{s}_k  \leq \boldsymbol{0},  
        \IEEEyessubnumber \label{eq:general nonlinear constraints} 
        \\
        \quad - \boldsymbol{s}_k \leq \boldsymbol{0},  \IEEEyessubnumber \label{eq:slack positivity}
        \\
        \quad\boldsymbol{x}_0 - \boldsymbol{x}(t) = \boldsymbol{0},  \quad \forall k \in \{0,1,\dots,N-1\}.
        \IEEEyessubnumber \label{eq:initial constraints}
        \end{IEEEeqnarray}
\end{Problem}
where $\boldsymbol{X} \triangleq \{\boldsymbol{x}_0, \dots, \boldsymbol{x}_{N}\}$ and $\boldsymbol{U} \triangleq \{\boldsymbol{u}_0, \dots, \boldsymbol{u}_{N-1}\}$ are the states and control variables along the control horizon N. The implicit discrete dynamics $\boldsymbol{f}_{\mathrm{impl}}(.):\mathbb{R}^n \times \mathbb{R}^n \times \mathbb{R}^m \mapsto \mathbb{R}^n$  \eqref{eq:discrete dynamics} captures the kino-dynamic equality path constraints in \eqref{eq:path integral constraints} discretized using first-order implicit-Euler integration scheme, and transcribed using Gauss-Legendre collocation method. The remaining nonlinear equality/inequality path constraints \eqref{eq:general nonlinear constraints} are implemented softly to avoid infeasibilities of the OCP by introducing extra slack variables $\boldsymbol{S} \triangleq \{\boldsymbol{s}_0, \dots, \boldsymbol{s}_{N}\}$, where $\boldsymbol{J}_{\mathrm{sh}}$ selects the slack variable attached to the respective constraint. The constraints $\boldsymbol{h}(.)$ are described in detail in \eqref{eq:kinodynamic constraint}-\eqref{eq:contact velocity constraint}. At every receding horizon, the initial condition of the OCP  is reset with the current measured state $\boldsymbol{x}(t)$ using constraint \eqref{eq:initial constraints}. We enforce Kino-dynamic consistency between the \eqref{eq:centroidal momentum dynamics}, and \eqref{eq:double integrator kinematics} with the following constraints:
\begin{IEEEeqnarray}{LLL} \label{eq:kinodynamic constraint}
    \boldsymbol{h}_{\mathrm{kindyn}} (\boldsymbol{c}_k, \boldsymbol{h}_{\mathcal{G}_k}, \tilde{\boldsymbol{q}}_k) + \boldsymbol{s}_{\mathrm{kindyn}_k} = \boldsymbol{0},\IEEEyesnumber\IEEEyessubnumber  
    \\
    \boldsymbol{h}_{\mathrm{kindyn}} \triangleq 
        \begin{bmatrix}
           \boldsymbol{c}_k - \textbf{COM}(\tilde{\boldsymbol{q}}_k), 
            \\
            \begin{bmatrix}
              \boldsymbol{\kappa}^\top_k, \, \boldsymbol{l}^\top_k    
            \end{bmatrix}^\top
             - \boldsymbol{A}_{\mathcal{G}} (\tilde{\boldsymbol{q}}_k) \dot{\boldsymbol{q}}_k 
        \end{bmatrix},  \IEEEyessubnumber 
\end{IEEEeqnarray} 
where $\textbf{COM}(.): \mathbb{Q} \mapsto \mathbb{R}^3$ function computes the center of mass of the robot for a given configuration, and $\boldsymbol{s}_{\text{kindyn}_k} \in \mathbb{R}^9$ are the slack variables associated with those kino-dynamic constraints. To avoid contact slippage, the tangential contact forces in the end-effector frame ($\boldsymbol{\mathfrak{f}}_{i,k} = \boldsymbol{R}^\top_{i,k} \boldsymbol{\lambda}_{i,k} $) are constrained inside the friction cone 
\begin{IEEEeqnarray}{lcr}  \label{eq:friction cone}
    \gamma_{i,k} \,.\, \left[h_{\mathrm{cone}_{i,k}} (\boldsymbol{\lambda}_{i, k}) + s_{\mathrm{cone}_{i, k}}\leq 0 \right]\quad \gamma_{i, k} \in \mathcal{C}, \IEEEyesnumber \IEEEyessubnumber 
    \\
    h_{\mathrm{cone}_{i, k}} \triangleq \sqrt{\mathfrak{f}^2_{{x,i}_k} + \mathfrak{f}^2_{{y, i}_k}} - \mu \mathfrak{f}_{{z, i}_k},  \IEEEyessubnumber
\end{IEEEeqnarray} 
where $\mathcal{C} = \{0, 1\}$. The contact mode (fixed apriori) $\gamma_{i,k} = 1$ when the $i$-th foot is in contact with the ground, and $\gamma_{i,k} = 0$ otherwise. The coefficient of friction is denoted by $\mu$, and $s_{\mathrm{cone}_{i, k}} \in \mathbb{R}$ is the slack variable associated to the friction cone constraint. During contact, the $i$-th end-effector position in the $z$-direction must be at the height of the contact surface $\mathcal{S}^z_{i,k}$, and be within the contact surface boundaries $\mathcal{S}^{x,y}_i$.  
\begin{IEEEeqnarray}{LLL} \label{eq:contact position constraint}
     \gamma_{i, k} \,.\, \left[h^z_{\mathrm{pos}_{i, k}} (\tilde{\boldsymbol{q}}_k) + s^z_{\mathrm{pos}_{i, k}} = \mathcal{S}^z_{i, k}\right] \quad \gamma_{i, k} \in \mathcal{C}, \IEEEyesnumber\IEEEyessubnumber \label{eq:contact location constraint z}
     \\
     \gamma_{i, k} \,.\, \left[\boldsymbol{h}^{x,y}_{\mathrm{pos}_{i, k}} (\tilde{\boldsymbol{q}}_k) + \boldsymbol{s}^{x,y}_{\mathrm{pos}_{i, k}} \in \mathcal{S}^{x,y}_{i, k}\right] \quad \gamma_{i, k} \in \mathcal{C}, \IEEEyessubnumber \label{eq:contact location constraint x-y} 
\end{IEEEeqnarray} 
where $\boldsymbol{h}_{\mathrm{pos}_{i, k}}\triangleq \text{FK}_{i, k}(\tilde{\boldsymbol{q}}_k) $, and $\boldsymbol{s}_{\mathrm{pos}_{i, k}} \in \mathbb{R}^3$ are the slack variables associated with the contact position constraints. For simplicity, we assume that $\mathcal{S}_{i,k} \in \mathbb{R}^3$ is a rectangular polytope. Finally, the end-effector velocities during contact are constrained to be zero by enforcing the holonomic~constraint:
\begin{IEEEeqnarray}{LLL} 
    \label{eq:contact velocity constraint} 
    \gamma_{i, k} \,.\, \left[\boldsymbol{h}_{\mathrm{vel}_{i,k}} (\tilde{\boldsymbol{q}}_k, \dot{\boldsymbol{q}}_k) + \boldsymbol{s}_{\mathrm{vel}_{i,k}} = \boldsymbol{0} \right] \quad \gamma_{i, k} \in \mathcal{C},  
\end{IEEEeqnarray} 
where $\boldsymbol{h}_{\mathrm{vel}_{i,k}} \triangleq \boldsymbol{J}_{i,k}(\tilde{\boldsymbol{q}}_k) \dot{\boldsymbol{q}}_k$, and $\boldsymbol{s}_{\mathrm{vel}_{i,k}} \in \mathbb{R}^3$ are the associated slack variables.
\subsection{Cost function and constraint penalties}
\label{sub-section:cost function}
In the above optimal control problem, we track a whole-body reference trajectory $\boldsymbol{x}_{\text{r}}$ optimized apriori offline. The total cost in \eqref{eq:nominal cost function} is split between the least-squares tracking cost $\mathcal{L}_{\text{LS}}$, and the penalty cost $\mathcal{L}_{\text{penalty}}$ penalizing the violations of the  nonlinear constraints \eqref{eq:general nonlinear constraints} as $\mathcal{L}_{\text{total}} = \mathcal{L}_{\text{LS}} + \mathcal{L}_{\text{penalty}}$
\begin{IEEEeqnarray}{LLL}
 \hspace{-2mm} \mathcal{L}_{\text{LS}}  \triangleq \hspace{-1mm}\sum^{N-1}_{k=0} 
    \hspace{-1mm} \frac{1}{2} \hspace{-1mm} \left(\norm{\boldsymbol{x}_k - \boldsymbol{x}_{\text{r}_k}}^2_{\boldsymbol{Q}}  + \norm{\boldsymbol{u}_k}_{\boldsymbol{R}}^2 \right)
    \hspace{-1mm} + \hspace{-1mm} \frac{1}{2}\norm{\boldsymbol{x}_N - \boldsymbol{x}_{\text{r}_N}}^2_{\boldsymbol{Q}_N} \IEEEyesnumber\IEEEyessubnumber \label{eq:least  squares cost}
    \\
    \mathcal{L}_{\text{penalty}} \triangleq  \mathlarger{\sum}^{N}_{k=0} \frac{1}{2} 
    \begin{bmatrix}
        \boldsymbol{s}_{l_k} \\ \boldsymbol{s}_{u_k} \\ \boldsymbol{1}        
    \end{bmatrix}^\top 
    \begin{bmatrix}
        \boldsymbol{Q}_{s_l}     & \boldsymbol{0}        & \boldsymbol{p}_{l} \\
        \quad\boldsymbol{0}  & \boldsymbol{Q}_{s_u}      & \boldsymbol{p}_{u} \\
        \boldsymbol{p}^\top_l& \boldsymbol{p}^\top_u & \boldsymbol{0}
    \end{bmatrix}
    \begin{bmatrix}
        \boldsymbol{s}_{l_k} \\ \boldsymbol{s}_{u_k} \\ \boldsymbol{1}        
    \end{bmatrix}. \IEEEyessubnumber
\end{IEEEeqnarray}
$\boldsymbol{Q} \in \mathbb{R}^{n \times n}$, and $\boldsymbol{R}^{m \times m}$ are the state and control running cost weight matrices respectively, while $\boldsymbol{Q}_N \in \mathbb{R}^{n \times n}$ is the terminal state cost weight. We assign both $l_1$ and $l_2$ penalties on the violations of the lower and upper bound nonlinear constraints \eqref{eq:general nonlinear constraints} associated with the slack variables $\boldsymbol{s}_{l_k}$ and $\boldsymbol{s}_{u_k}$ respectively, where $\boldsymbol{p}_l$, $\boldsymbol{p}_u$ are the $l_1$ penalty weights, and $\boldsymbol{Q}_{s_l}$, $\boldsymbol{Q}_{s_u}$ are the $l_2$ penalty weights. Notice that the slack variables are constrained to be positive in \eqref{eq:slack positivity} to attain the effect of an $l_1$ penalty as explained in~\cite{nocedal1999}.      
\section{Stochastic Optimal Control for Kinodynamic Trajectory Optimization}
\label{subsection:stochastic NOCP}
We present the stochastic OCP version of problem \eqref{NMPC problem}. Consider the following controlled stochastic diffusion:
\begin{IEEEeqnarray}{LLL}
\label{eq:controlled stochastic diffusion}
\partial\boldsymbol{x} = \partial\boldsymbol{f}(\boldsymbol{x}_t, \boldsymbol{u}_t)\partial t + \boldsymbol{C} \partial\boldsymbol{w}_t, 
\end{IEEEeqnarray}
where the state $\boldsymbol{x}_t \in \mathbb{R}^{n}$ evolves stochastically based on the additive random variable $\boldsymbol{w}_t \in \mathbb{R}^{n}$. The selector matrix $\boldsymbol{C} \in \mathbb{R}^{n \times n}$ maps the additive disturbance on the dynamics. 
\begin{assumption}(Additive disturbance process)
\\
\label{assumption:gaussian noise assumption}
 $\partial\boldsymbol{w}_t \sim \mathcal{N}(\boldsymbol{0}, \partial t)$ is an additive Gaussian random process with zero mean.  
\end{assumption}
%
\begin{assumption}(State feedback control policy) \\
\label{assumption:state feedback control policy}
$\boldsymbol{u}_t \in \mathbb{R}^{m}$ is a causal state feedback control policy in the form of $\boldsymbol{u}_t = \boldsymbol{u}^\ast_t + \boldsymbol{K} (\boldsymbol{x}^\ast_t - \boldsymbol{x}_t)$, where $\boldsymbol{u}^\ast_t$ is the optimized feedforward open-loop control actions,  $\boldsymbol{K} \in \mathbb{R}^{m \times n}$ are stabilizing feedback gains, and $\boldsymbol{x}_t$ is the deterministic state evolving as $\boldsymbol{x}_{t+1} = \boldsymbol{f}(\boldsymbol{x}_t, \boldsymbol{u}_t)$.   
\end{assumption}
Given the above, we aim to solve the following SNMPC problem with contact location chance constraints.
\begin{Problem} Kino-dynamic SNMPC problem with contact location joint chance-constraints
\label{problem:SOCP}
    \begin{IEEEeqnarray}{LLL}
        \minimize_{\substack{\boldsymbol{X}, \boldsymbol{U}, \boldsymbol{S}}} \,\,  \mathbb{E} \left[ \mathcal{L}_{\mathrm{total}}(\boldsymbol{X}, \boldsymbol{U}, \boldsymbol{S}) \right]\IEEEyesnumber\IEEEyessubnumber \label{eq:expected cost function} 
        \\
       \,\, \mathrm{s.t.} \nonumber 
       \\
       \qquad\boldsymbol{f}^\prime\left(\boldsymbol{x}_{k+1}, \boldsymbol{x}_k, \boldsymbol{u}_k, \boldsymbol{w}_k\right) =  \boldsymbol{0} \IEEEyessubnumber, \label{eq:discrete stochastic dynamics}
        \\
        \qquad \eqref{eq:kinodynamic constraint},  \eqref{eq:friction cone}, \eqref{eq:contact location constraint z}, \eqref{eq:contact velocity constraint}, \IEEEyessubnumber \label{eq:deterministic constraints}
        \\
        \qquad \mathrm{Pr} \eqref{eq:contact location constraint x-y} \geq \alpha_{i, k}, \quad \forall i \in \{1,\dots,n_c\}, \IEEEyessubnumber \label{eq:contact position chance-constraint}
        \\
        \qquad -\boldsymbol{s}_k \leq \boldsymbol{0},  \qquad \,\,\, \forall k \in \{0,1,\dots,N-1\},\IEEEyessubnumber
        \\
        \qquad\boldsymbol{x}_0 - \boldsymbol{x}(t) = \boldsymbol{0},  \IEEEyessubnumber
    \end{IEEEeqnarray}
\end{Problem}
where \eqref{eq:discrete stochastic dynamics} are the discrete-time implicit stochastic dynamics equality path constraints in \eqref{eq:controlled stochastic diffusion}. The constraints \eqref{eq:deterministic constraints} are to be satisfied deterministically by enforcing them on the mean of the state. In this work, we aim to account for the additive uncertainties by enforcing the contact location constraints in the $x-y$ directions probabilistically within at least a probability level $\alpha_{i, k}$ \eqref{eq:contact position chance-constraint}, which are known as chance-constraints. The above SNMPC problem is not tractable in general since the dynamics are stochastic. Moreover, resolving the chance-constraints in \eqref{eq:contact position chance-constraint} requires the integration of multi-dimensional Probability Density Functions (PDFs), which becomes computationally intractable for high dimensions. To tackle those issues, we solve an approximate deterministic reformulation of the above OCP.
\vspace{-0.1 cm}  

    \newtheorem{property}{Property}
\newtheorem{remark}{Remark}
\subsection{Tractable formulation of Joint Chance-constraints}
The goal of the following subsections is to design safety margins known as \textit{back-offs} on the contact location chance-constraints \eqref{eq:contact position chance-constraint} to account for the additive stochastic disturbances on the dynamics that is difficult for only feedback to deal with. This is crucial for legged robots since making contact at unintended contact locations can lead to unsafe motions. The back-offs magnitudes are designed based on the evolution of statistical information along the horizon inside the OCP, such that we can provide probabilistic statements about constraint satisfaction without degrading the performance. Notice that those margins are not fixed compared to designing them heuristically by hand (check the results section). In other words, if the variance of the uncertainty is large, or we want to satisfy the constraints with larger probability, then it reflects automatically on the back-off magnitude by increasing the safety margin accordingly to ensure the expected probability of constraint satisfaction. To reduce the computational complexity for solving the contact-location joint chance-constraints \eqref{eq:contact position chance-constraint}, we first linearize the nonlinear constraints around the $j$-th SQP iteration at $\Delta\boldsymbol{x}_k \triangleq \boldsymbol{x}_k - \boldsymbol{x}^j_k$, then solve for each individual half-space chance-constraints forming the linearized feasible set as: 
\begin{IEEEeqnarray}{LLL}
    \label{eq: linearized contact location chance-constraints}
  \hspace{-6mm}  \nabla_{\boldsymbol{x}_k} \eqref{eq:contact location constraint x-y} &= \gamma_{i,k} . \Big(\boldsymbol{h}^{x,y}_{\mathrm{pos}_{i, k}}(\tilde{\boldsymbol{q}}^j_{k}) + \boldsymbol{J}^{x,y}_{i, k} (\tilde{\boldsymbol{q}}^j_k) \Delta\boldsymbol{x}_k \in \mathcal{S}^{x,y}_{i,k}\Big). \IEEEyesnumber
\end{IEEEeqnarray}
Given the above linearized constraints, we  write down the contact location chance constraints as a conjunction of half-space  constraints of the 2D polygon forming $\mathcal{S}^{x, y}_{i,k} \in \mathbb{R}^4$ as
\begin{IEEEeqnarray}{LLL}
\label{eq:disjunctive contact location chance-constraints}
    \mathrm{Pr}\left(\bigwedge^{4}_{l=1} \boldsymbol{G}_{i, k}^{l}(\tilde{\boldsymbol{q}}^j_{k}) \boldsymbol{x}_k + g^l_{i, k} \in \mathcal{S}^{x, y}_{i,k}\right) \geq \alpha_{i,k} \nonumber 
    \\
    \equiv 
    \mathrm{Pr}\left(\bigvee^{4}_{l=1}  \boldsymbol{G}^{l}_{i, k}(\tilde{\boldsymbol{q}}^j_{k}) \boldsymbol{x}_k + g^l_{i, k} \notin \mathcal{S}^{x,y} \right) \hspace{-1mm}\leq 1 - \alpha_{i,k}, \IEEEyesnumber
\end{IEEEeqnarray}
where $\boldsymbol{G}^{l}_{i, k}$ is the $l$-th row of $\pm \boldsymbol{J}^{x,y}_{i,k}(\tilde{\boldsymbol{q}}^j_k) \in \mathbb{R}^{4 \times n}$, and $g^l_{i, k}$ is the $l$-th element of the vector $\boldsymbol{g}_{i,k} \pm\left(\boldsymbol{J}^{x,y}_{i, k} \boldsymbol{x}^j_k - \boldsymbol{h}^{x,y}_{\mathrm{pos}_{i, k}}\right)$. To avoid multi-dimensional integrals of joint chance constraints, we use Boole's inequality 
\begin{IEEEeqnarray}{LLL}
\label{eq:boole's inequality}
    \text{Pr}(\bigvee^n_{i=1} \boldsymbol{C}_i)\leq \sum^n_i \text{Pr} (\boldsymbol{C}_i), 
\end{IEEEeqnarray}
as a conservative union bound on the joint chance constraints \cite{Ono2010}. By applying Boole's inequality we get
\begin{IEEEeqnarray}{LLL}
    \label{eq:indvidual contact location chance constraints}
    \eqref{eq:disjunctive contact location chance-constraints}  & \xLeftarrow[]{(\ref{eq:boole's inequality})} \sum^4_{l=1}\mathrm{Pr}\left(\boldsymbol{G}^{l}_{i, k}(\tilde{\boldsymbol{q}}^j_{k}) \boldsymbol{x}_k > g^l_{i, k}\right) \leq 1-\alpha_{i,k}.  
\end{IEEEeqnarray}
By allocating equal risk for constraint violation for each half-space constraint $\epsilon_{i,k} \triangleq (1-\alpha_{i,k})/4$, we finally arrive to:
\begin{IEEEeqnarray}{LLL}
    &\eqref{eq:indvidual contact location chance constraints} \xLeftarrow[] {} \mathrm{Pr}\left( \boldsymbol{G}^{l}_{i, k}(\tilde{\boldsymbol{q}}^j_{k}) \boldsymbol{x}_k +  g^l_{i, k} \notin \mathcal{S}^{x,y}\right) \leq  \epsilon_{i,k},  \nonumber
    \\
    &\equiv \mathrm{Pr}\left(\boldsymbol{G}^{l}_{i, k}(\tilde{\boldsymbol{q}}^j_{k}) \boldsymbol{x}_k +  g^l_{i, k} \in \mathcal{S}^{x,y}\right) \geq  1-\epsilon_{i,k}, \nonumber
    \\
    &\forall{i=1,\dots,n_c}, \,\, \forall{k=0,\dots,N-1}, \,\, \forall{l=1,\dots,4}.\label{eq:indvidual contact location chance constraints final} 
\end{IEEEeqnarray}
Given the above linearized chance constraints, we proceed to solve for each unimodal PDF representing the half-space chance constraints forming the contact surface polygon. 
\subsection{Deterministic Reformulation of Individual Contact location Chance-Constraints}
\begin{figure}[!t]
    \includegraphics[scale=0.3]{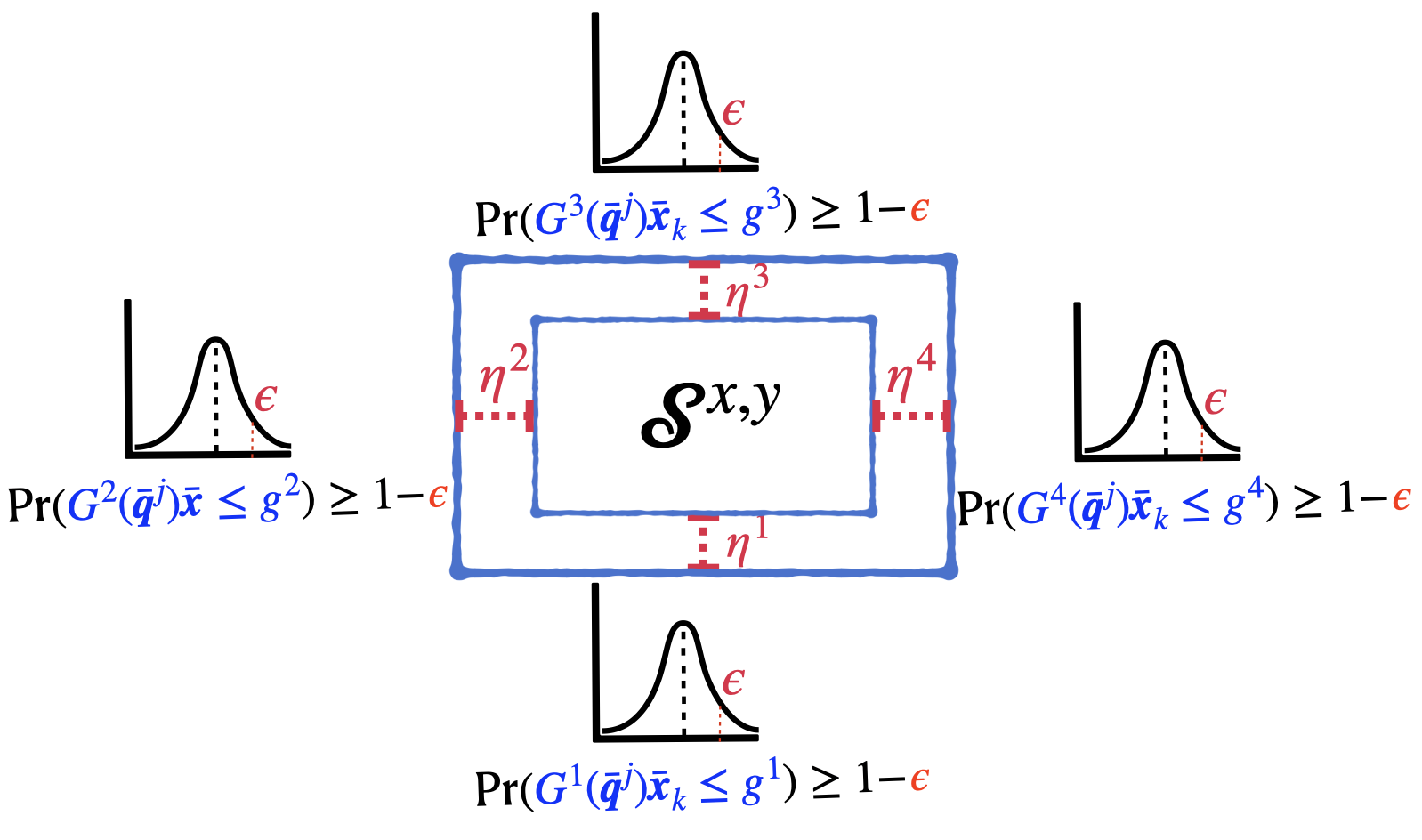}
    \caption{Effect of equally distributed back-offs design of the linearized contact location chance-constraints.}
    \label{fig:back-offs}
\end{figure}

In this subsection, we present a tractable deterministic formulation for solving the above individual chance constraints \eqref{eq:indvidual contact location chance constraints final}. This requires statistical knowledge of the uncertainty propagation through nonlinear dynamics. One way to do this is by exploiting sampling methods like unscented-based transforms \cite{plancher2017}, or Generalized Polynomial Chaos (gPC) \cite{nakka2019}. These methods can lead to a more accurate estimate of the propagated uncertainty at a cost of a significant increase in computational complexity, which is time demanding for real-time MPC. Since computational efficiency is critical in our case, we defer from using these methods. Instead, we adopt an approximate linearization-based covariance propagation as in \cite{zhu2019}\cite{lew2020}. Based on assumptions \eqref{assumption:gaussian noise assumption}-\eqref{assumption:state feedback control policy}, the approximate mean and covariance of the dynamics evolve as 
\begin{IEEEeqnarray}{LLL}
    \bar{\boldsymbol{x}}_{k+1} & \approx \boldsymbol{f}(\bar{\boldsymbol{x}}_k, \bar{\boldsymbol{u}}_k) + \boldsymbol{A}_k \Delta \bar{\boldsymbol{x}}_k  + \boldsymbol{B}_k \Delta \bar{\boldsymbol{u}}_k, \IEEEyesnumber\IEEEyessubnumber\label{eq:linearized dynamics}
    \\
    \boldsymbol{\Sigma}_{\boldsymbol{x}_{k+1}} &=  \boldsymbol{A}_{\text{cl}}\boldsymbol{\Sigma}_{\boldsymbol{x}_k}  \boldsymbol{A}_{\text{cl}}^\top + \boldsymbol{C} \boldsymbol{\Sigma_w}, \quad \boldsymbol{\Sigma}_{\boldsymbol{x}_0} = \boldsymbol{0}, \IEEEyessubnumber \label{eq:covariance propagation}
    \\
    \boldsymbol{A}_\text{cl} &\triangleq \boldsymbol{A}_k + \boldsymbol{B}_k \boldsymbol{K}_k. \IEEEyessubnumber\label{eq:closed-loop dynamics}
\end{IEEEeqnarray}
The initial condition is assumed to be deterministic (i.e. $\boldsymbol{x}_0 = \bar{\boldsymbol{x}}_0$), where the $\bar{}$ superscript denotes the mean of the quantity. The $j$-th controls perturbation is defined as $\Delta \bar{\boldsymbol{u}}_k \triangleq \bar{\boldsymbol{u}}_k - \bar{\boldsymbol{u}}^j_k$, and \eqref{eq:closed-loop dynamics} is the linearized closed-loop dynamics, where $\boldsymbol{A}_k \triangleq \nabla_{\bar{\boldsymbol{x}}} \boldsymbol{f}(\bar{\boldsymbol{x}}_k, \bar{\boldsymbol{u}}_k)|_{\bar{\boldsymbol{x}}^j_k,  \bar{\boldsymbol{u}}^j_k}$, and $\boldsymbol{B}_k \triangleq \nabla_{\bar{\boldsymbol{u}}} \boldsymbol{f}(\bar{\boldsymbol{x}}_k, \bar{\boldsymbol{u}}_k)|_{\bar{\boldsymbol{x}}^j_k,  \bar{\boldsymbol{u}}^j_k}$ are the Jacobians of the dynamics w.r.t. the nominal state and control respectively. The feedback gains $\boldsymbol{K}_k$ are computed using the Discrete Algebraic Riccati Equation (DARE) based on the time-varying linearized dynamics.
Using the covariance information in (\ref{eq:covariance propagation}) and the probability level of constraint violation $\epsilon_{i,k}$, we seek the least conservative upper bound $\eta^l_{i,k}$ at each point in time for each half-space contact location chance-constraint \eqref{eq:indvidual contact location chance constraints final}. By exploiting the inverse of the \textit{Cumulative Density Function} (CDF) $\Phi^{-1}$ of a Gaussian distribution, we get a deterministic reformulation of the individual contact location chance-constraints:
\begin{IEEEeqnarray}{LLL}
    \label{eq:chance-constraints backoffs}
    \IEEEyesnumber
    \boldsymbol{G}^{l}_{i, k}(\tilde{\boldsymbol{q}}^j_{k}) \bar{\boldsymbol{x}}_k + g^l_k \in \mathcal{S}^{x,y} - \eta^l_k,  \quad \forall l = {1, \dots, 4},\IEEEyesnumber\IEEEyessubnumber 
    \\
    \eta^l_k \triangleq \Phi^{-1}(1-\epsilon_{i,k}) \norm{\boldsymbol{G}^{l}_i}_{\boldsymbol{\Sigma}_{\boldsymbol{x}_k}} \IEEEyessubnumber. 
\end{IEEEeqnarray}
$\eta^l_k$ is the back-off bound that ensures the satisfaction of the individual chance constraints \eqref{eq:indvidual contact location chance constraints final} illustrated pictorially in Fig.~\ref{fig:back-offs}. Contrary to designing a heuristic-based back-off bound by hand, this upper bound is not fixed (i.e. it varies at every point in time along the horizon). This is because the magnitude of this back-off bound is scaled by the covariance propagation along the horizon \eqref{eq:covariance propagation}, the magnitude of the time-varying feedback gain $\mathbf{K}_k$~\eqref{eq:closed-loop dynamics}, as well as the design parameter $\epsilon_{i, k}$ capturing the desired probability of constraint satisfaction. Although in theory, one can optimize for both $\boldsymbol{K}_k$ and $\epsilon_{i, k}$ for better performance, this usually leads to a bi-level optimization, which is computationally expensive to solve in real-time for MPC applications.
\subsection{Deterministic Reformulation of SNMPC}
Given the previous reformulation of the individual chance constraints, we can write down a deterministic reformulation of the original SNMPC problem \eqref{problem:SOCP} on the mean of the nonlinear dynamics. Despite the reformulated chance-constraints constraints, this problem has the same number of optimization variables as the NMPC problem \eqref{NMPC problem}, which means that with the gained robustness, we don't increase the computational complexity of the problem over NMPC.  
\begin{Problem}
\label{problem:CCOCP}
Kino-dynamic SNMPC problem with individual chance-constraints
    \begin{IEEEeqnarray}{LLL}
        \minimize_{\substack{\bar{\boldsymbol{X}}, \bar{\boldsymbol{U}}, \boldsymbol{S}}} \,\, \mathcal{L}_{\mathrm{total}}(\bar{\boldsymbol{X}}, \bar{\boldsymbol{U}}, \boldsymbol{S}) \IEEEyesnumber\IEEEyessubnumber 
        \\
        \,\,\mathrm{s.t.} \,\, \boldsymbol{F}(\bar{\boldsymbol{x}}_{k+1},\bar{\boldsymbol{x}}_k, \bar{\boldsymbol{u}}_k) \triangleq  \boldsymbol{f}_{\mathrm{impl}}(\bar{\boldsymbol{x}}_k, \bar{\boldsymbol{x}}_{k+1}, \bar{\boldsymbol{u}}_k) = \boldsymbol{0},  \IEEEyessubnumber
        \\
        \qquad \boldsymbol{E}(\bar{\boldsymbol{x}}_k, \bar{\boldsymbol{u}}_k) \triangleq
            \left\{ 
                \begin{array}{ll}
                    \boldsymbol{h}_{\mathrm{eq}}(\bar{\boldsymbol{x}}_k, \bar{\boldsymbol{u}}_k, \boldsymbol{s}_k) = \boldsymbol{0}, 
                     \\
                     \bar{\boldsymbol{x}}_0 -\boldsymbol{x}(t) = \boldsymbol{0},          
                \end{array}
            \right. \IEEEyessubnumber \label{eq:equality constraints}
        \\
        \qquad \boldsymbol{I}(\bar{\boldsymbol{x}}_k, \bar{\boldsymbol{u}}_k) \triangleq 
            \left\{
                \begin{array}{ll}
                    \boldsymbol{h}_{\mathrm{ineq}}(\bar{\boldsymbol{x}}_k, \bar{\boldsymbol{u}}_k, \boldsymbol{s}_k) \leq \boldsymbol{0},   
                    \\
                    \boldsymbol{G}(\bar{\boldsymbol{x}}^j_{k}) \bar{\boldsymbol{x}}_k + \boldsymbol{g}_k + \boldsymbol{J}_{\mathrm{sg}} \boldsymbol{s}_k \in \mathcal{S}^{x,y}  - \boldsymbol{\eta}_k,  
                \end{array} 
            \right. \IEEEyessubnumber \label{eq:inequality constraints}
        \\
        \qquad -\boldsymbol{s}_k \leq \boldsymbol{0}, \quad \forall k \in \{0,1,\dots,N-1\}. \IEEEyessubnumber
    \end{IEEEeqnarray}
\end{Problem}      
This SMPC problem optimizes for the open-loop mean states $\bar{\boldsymbol{X}} = \{\bar{\boldsymbol{x}}_0, \dots, \bar{\boldsymbol{x}}_N\}$, and feedforward controls $\bar{\boldsymbol{U}} = \{\bar{\boldsymbol{u}}_0, \dots, \bar{\boldsymbol{u}}_N\}$. All nonlinear equality constraints are captured inside $\boldsymbol{h}_{\mathrm{eq}}(.) = \small \begin{bmatrix} \eqref{eq:kinodynamic constraint}^\top, \eqref{eq:contact location constraint z}^\top,  \eqref{eq:contact velocity constraint}^\top \end{bmatrix}^\top$, while $\boldsymbol{h}_{\mathrm{ineq}} (.) = \eqref{eq:friction cone}$ captures the friction cone inequality constraints. Finally, the second row of the inequality constraints \eqref{eq:inequality constraints} are the backed-off contact location constraints in the $x-y$ directions, where $\boldsymbol{\eta_k} = \eta_k.\boldsymbol{1}_{2n_c}$. These linearized constraints are implemented softly with $\boldsymbol{J}_\mathrm{sg}$ being the slack selector matrix. The above OCP is solved using Sequential Quadratic Programming (SQP)~\cite{nocedal1999} by constructing a quadratic model of the cost objective subject to linearized constraints that solves the Karush-Kuhn-Tucker (KKT) system associated with the following Lagrangian:
\begin{IEEEeqnarray}{LLL} \label{eq:lagrangian}
\Psi(\boldsymbol{z}) = \mathcal{L}_\mathrm{total} + \boldsymbol{\zeta}^\top \boldsymbol{F} + \boldsymbol{\beta}^\top \boldsymbol{E} + \boldsymbol{\gamma}^\top  \boldsymbol{I}, \IEEEyesnumber
\end{IEEEeqnarray}
where $\boldsymbol{z} \triangleq [\bar{\boldsymbol{x}}^\top, \bar{\boldsymbol{u}}^\top]^\top$ is the concatenated vector of states and controls. $\boldsymbol{\zeta}$, $\boldsymbol{\beta}$ are the associated Lagrange multipliers to the equality constraints, and $\boldsymbol{\gamma}$ are the ones corresponding to the inequality constraints. Given a perturbation $\Delta \boldsymbol{z}_k \triangleq \boldsymbol{z}_k - \boldsymbol{z}^j_k$, where $\boldsymbol{z}^j$ is the current initial guess along the control horizon, the following QP subproblem is solved: 
\begin{algorithm}[!t]
\DontPrintSemicolon
 $\boldsymbol{Z}^0_{\mathrm{traj}} \gets$ \text{Offline whole-body iLQR initial guess}\;
\For{$j = 0, 1, \dots, N_{\mathrm{traj}-1} \,(\mathrm{trajectory\, horizon})$}{
\tcc{QP PREPARATION PHASE}
$ \boldsymbol{Z}^j \gets \textbf{initial guess}\,\, 
 \begin{cases}
                     \boldsymbol{Z}^0_{\mathrm{traj}} \quad   \text{if} \,\,j = 0 \\
                     \boldsymbol{Z}^{j-1}_{\mathrm{mpc}} \quad \text{otherwise}
\end{cases}$\;
\For{$k = 0, 1, \dots, N_{\mathrm{mpc}}-1 \, (\mathrm{MPC\, horizon})$}{
$\boldsymbol{A}^j_k, \boldsymbol{B}^j_k \gets \mathrm{computeFWDSensitivities} ~\eqref{eq:linearized dynamics}$\;
$\boldsymbol{K}^j_k \gets \mathrm{computeDARE} (\boldsymbol{A}^j_k, \boldsymbol{B}^j_k, \boldsymbol{Q}, \boldsymbol{R})$\;
$\boldsymbol{\Sigma}^j_k \gets \mathrm{propagateCovariance}~\eqref{eq:covariance propagation}-\eqref{eq:closed-loop dynamics}$\;
$\boldsymbol{\eta}^j_k \gets \mathrm{computeBackOffs}~\eqref{eq:chance-constraints backoffs}$}
\tcc{FEEDBACK PHASE}
$\boldsymbol{\Delta \boldsymbol{Z}^\ast}, \boldsymbol{S}^\ast \gets \mathrm{solveProblem}~\ref{problem:QP}$\;
$\boldsymbol{u}^j_0 \gets \mathrm{ApplyControlPolicyAssumption} ~\ref{assumption:state feedback control policy}$\;                
$\mathbf{Z}^{j+1} \gets \mathbf{Z}^j + \Delta \mathbf{Z}^\ast, \,\, \boldsymbol{\zeta}^{j+1} = \boldsymbol{\zeta}^\ast$\;
$\boldsymbol{\beta}^{j+1} \gets \boldsymbol{\beta}^\ast, \,\, \boldsymbol{\gamma}^{j+1} \gets \boldsymbol{\gamma}^\ast $\;
}
\KwOutput{$\textbf{X}^\ast_{\mathrm{traj}},  \textbf{U}^\ast_{\mathrm{traj}}, \textbf{S}^\ast_{\mathrm{traj}}$}
\caption{Approximate SMPC Algorithm.}
\label{alg:SMPC}
\end{algorithm}
\setlength{\textfloatsep}{0pt}
\begin{Problem}
\label{problem:QP}
QP subproblem
    \begin{IEEEeqnarray}{LLL}
        \minimize_{\substack{ \Delta\boldsymbol{Z}, \boldsymbol{S}}} \,\, \frac{1}{2} \Delta \boldsymbol{z}^{j^\top} \boldsymbol{H} \Delta \boldsymbol{z}^j + \boldsymbol{p}^\top \Delta \boldsymbol{z}^j \IEEEyesnumber\IEEEyessubnumber 
        \\
        \,\,\mathrm{s.t.} 
        \,\,\, \boldsymbol{F}(\boldsymbol{z}^j) + \nabla_{\boldsymbol{z}}   \boldsymbol{F} (\boldsymbol{z}^j) \Delta \boldsymbol{z}^j = \boldsymbol{0},\IEEEyessubnumber  \label{eq: linearized ddynamics constraints}
        \\
        \qquad \boldsymbol{E}(\boldsymbol{z}^j) + \nabla_{\boldsymbol{z}} \boldsymbol{E} (\boldsymbol{z}^j) \Delta \boldsymbol{z}^j = \boldsymbol{0}, \IEEEyessubnumber \label{eq: linearized equality constraints}
        \\
        \qquad \boldsymbol{I}(\boldsymbol{z}^j) \,\,+ \nabla_{\boldsymbol{z}} \boldsymbol{I} (\boldsymbol{z}^j) \Delta \boldsymbol{z}^j \leq \boldsymbol{0}, \,\,\, \boldsymbol{s}^j \leq \boldsymbol{0}  \IEEEyessubnumber \label{eq: linearized inequality constraints}.
    \end{IEEEeqnarray}
\end{Problem}    
The Hessian of the Lagrangian $\boldsymbol{H} \triangleq \nabla^2_{\boldsymbol{z}} \Psi(\boldsymbol{z}^j)$ is approximated using the Generalized Gauss-Newton (GGN) variant of SQP as $\boldsymbol{H} \approx \nabla^\top_{\boldsymbol{z}}\Psi(\boldsymbol{z}^j) \nabla_{\boldsymbol{z}} \Psi(\boldsymbol{z}^j)$, and the gradient of the residual is defined as $\boldsymbol{p} \triangleq \nabla^\top_{\boldsymbol{z}} \Psi(\boldsymbol{z}^j) \Psi(\boldsymbol{z}^j)$ \cite{diehl2019}. For an exact SQP iteration, the linearization of the backed-off contact location constraints included in the above inequality constraints includes the extra derivative $\nabla_{\boldsymbol{z}}\eta(\boldsymbol{z}^j_k)$:
\begin{IEEEeqnarray}{LLL}
    \underbrace{\boldsymbol{G}^{l}_{i, k}(\bar{\boldsymbol{x}}^j_{k}) \bar{\boldsymbol{x}}_k \leq g^l_{i,k} - \eta^l_{k}}_{\boldsymbol{I}_{\mathrm{\boldsymbol{G}}}(\boldsymbol{z}^j_k)} - \nonumber
    \\  
    \qquad \underbrace{\Phi^{-1}(1- \epsilon^l_{i,k})\Big( \nabla_{\boldsymbol{z}}\norm{\boldsymbol{G}^{l}_{i,k}  (\bar{\boldsymbol{x}}^j_{k})}_{\boldsymbol{\Sigma}_{\boldsymbol{x}_k}}}_{\nabla_{\boldsymbol{z}}\eta(\boldsymbol{z}^j_k)}(\boldsymbol{z}_k -\boldsymbol{z}_k^j)\Big), \nonumber 
    \\
    \nabla_{\boldsymbol{z}}\norm{\boldsymbol{G}^{l}_{i,k}  (\bar{\boldsymbol{x}}^j_{k})}_{\boldsymbol{\Sigma}_{\boldsymbol{x}_k}} \triangleq  \big(2\norm{\boldsymbol{G}^l_{i, k}}_{\boldsymbol{\Sigma}_{\boldsymbol{x}_k}}\big)^{-1}\Big(2\boldsymbol{G}^{l^\top}_{i,k} \boldsymbol{\Sigma}_{\boldsymbol{x}_k} \nabla_{\boldsymbol{z}}\boldsymbol{G}^l_{i,k} \hspace{-0.07cm}+ \nonumber 
    \\ 
    \qquad\qquad\qquad\qquad\qquad\quad \sum^n_{i=0} \sum^n_{j=0} g^l_i g^l_j \nabla_{\boldsymbol{z}} \Sigma_{ij}\Big), \IEEEyesnumber\IEEEyessubnumber
    \\ \forall l \in \{1, \dots, 4\}, \forall i \in \{1, \dots, n_c\}, \forall k \in \{0, \dots, N\}. \IEEEyessubnumber
\end{IEEEeqnarray}
The above derivative involves the tensor derivative of the covariance matrix $\nabla_{\boldsymbol{z}} \boldsymbol{\Sigma}_{\boldsymbol{x}_k}$, which is expensive to compute. 
\begin{remark} 
For real-time computational tractability,  we adopt a SQP-type iteration by approximating $\nabla_{\boldsymbol{z}} \eta(\boldsymbol{z}^j_k) = 0$ as in ~\cite{hewing2018cautious}. This SQP approach is sub-optimal since we don't compute the exact Jacobian of the contact location inequality constraint as in ~\cite{gazar2023},~\cite{lew2020}. Despite this sub-optimality, this scheme yields good results in practice without sacrificing computational complexity over NMPC.
\end{remark}
The OCP is implemented with real-time iteration~\cite{diehl2005real}, where one QP sub-problem is solved at a time using a full Newton-type step without a line search (see Algorithm~\ref{alg:SMPC}).

\section{Simulations Results}
\label{sec:results}
We report simulation results comparing SNMPC against NMPC for the quadruped robot Solo~\cite{grimminger2020open} performing dynamic trotting and bounding gaits on non-coplanar small stepping stones. The robustness of both controllers is tested in terms of contact location constraint satisfaction (safety), and performance computed using the least-squares tracking cost \eqref{eq:least squares cost}. Moreover, we tested the safety margins induced by SNMPC against heuristic-based NMPC (HNMPC) and NMPC. For HNMPC, we shrank the contact-location constraints heuristically by hand by performing a grid search on an interval between 1 cm and 3 cm. A safety margin of 3 cm was selected as it was the first value where the contact-location constraints became active and differed from NMPC. 
\\ 
\indent We conducted two sets of simulations: A) \textbf{Kino-dynamic Monte-Carlo Simulations}, where we test the robustness of the kino-dynamic model against persistent disturbance realizations. B) \textbf{Whole-body simulations}, to test the effect of model mismatch between the kino-dynamic model and whole-body model of the robot in the Pybullet simulator. The three MPC controllers follow a generated offline using whole-body DDP from the Croccodyl solver ~\cite{mastalli2020crocoddyl} with pre-planned contact locations at the center of the contact surfaces. Also, the first MPC iteration is warm-started using this trajectory, while subsequent MPC iterations are warm-started from the previous MPC solution. 
All problems were discretized with a sampling time of $\Delta_k = 10$ ms for an MPC horizon length of $N = 40$, and $N = 55$ for the trot and bound motions respectively. The motion plans were designed with a coefficient of friction $\mu = 0.5$. Finally, the real-time iteration scheme was performed using the optimal control solver ACADOS~\cite{Verschueren2021}, exploiting Casadi's automatic differentiation~\cite{Andersson2019}, and Pinocchio's analytical derivatives for rigid body kinematic functions for computing the  underlying derivatives~\cite{jain1993l, carpentier2018analytical, singh2022}.
\begin{table}[t!]
    \caption{Robustness and performance.}
    \begin{subtable}[c]{0.45\columnwidth}
        \caption{Rate of successful motions}
        \setlength\tabcolsep{1.5pt}
        \label{table:robustness}
        \begin{tabular}{cccc}
            \toprule 
            \textbf{Task} &{NMPC} & {HNMPC} & {SNMPC} 
            \\
            \midrule
            Trot & 51.0\%    &85.4\% &\textbf{100}\%
            \\
            Bound &52.4\%  &67\%  &\textbf{100}\%
            \\
            \bottomrule
        \end{tabular} 
    \end{subtable}
    \setlength\tabcolsep{1.5pt}
    \begin{subtable}[c]{0.45\columnwidth}
        \caption{Open-loop MPC cost.}
        \label{table:performance}
        \begin{tabular}{cccc}
            \toprule 
            \textbf{Task} &{NMPC} & {HNMPC} &{SNMPC} 
            \\
            \midrule
            Trot$\times 10^7$  & $\mathbf{3.27}$ &   $3.28$ & $3.31$
            \\
            Bound$\times 10^8$ & $\mathbf{4.72}$ &  $\mathbf{4.72}$  & $4.77$
            \\
            \bottomrule
        \end{tabular}
    \end{subtable}
    \label{table:robustness and performance}
\end{table}
\begin{figure}[t!]
    \centering 
    \begin{subfigure}[t]{.45\columnwidth}
        \centering 
        \includegraphics[scale=0.28]{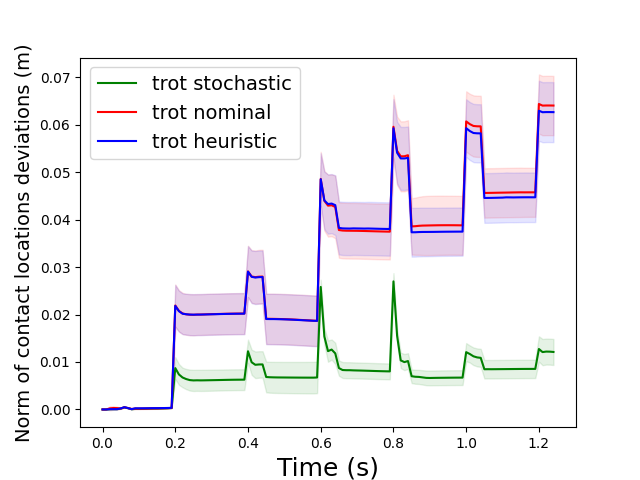}
        \caption{Trot motion.}
    \end{subfigure}
    \begin{subfigure}[t]{.45\columnwidth}
        \centering
        \includegraphics[scale=0.28]{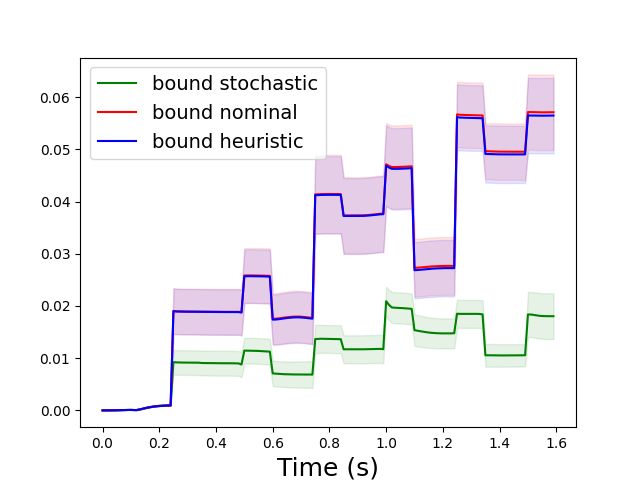}
        \caption{Bound motion.}
    \end{subfigure}
    \caption{Norm of the contact location deviations from the contact surface center using NMPC, HNMPC, and SNMPC.} 
    \label{fig:contact-location deviations norm}
\end{figure}
\subsection{Kino-dynamic Monte-Carlo Simulations}
We run 500 closed-loop kino-dynamic Monte-Carlo simulations for each motion (trotting and bounding). We sample additive kinematic disturbance realizations from a multi-variate Gaussian distribution with zero mean and a covariance $\boldsymbol{\Sigma}_w = \textbf{DIAG}\,\small[\boldsymbol{0}_6, 0.3^2, 0.3^2, 0.3^2, 0.2^2, 0.2^2, 0.2^2, 0.7^2, 0.7^2, 0.7^2, 0.7^2, 
\\
0.7^2, 0.7^2, 0.7^2, 0.7^2, 0.7^2, 0.8^2, 0.8^2, 0.8^2, 0.1^2, 0.1^2, 0.1^2, 0.7^2, \\
0.7^2,0.7^2, 0.7^2, 0.7^2, 0.7^2, 0.7^2, 0.7^2, 0.7^2]^\top$. We tune the risk of violating the contact location constraints for SNMPC to be $\epsilon = 0.01$ for all feet and contact surfaces. The disturbances are applied on the base velocity at the time 
contacts are made to mimic the effect of impacts on the kino-dynamic model, as well as on the swing leg joint velocities during take-off and landing to simulate persistent disturbances and control imperfections on the swing legs. Finally, no disturbances are applied at the feet after impact based on the assumption that the feet do not slip. The disturbance realizations are discretized and integrated on the dynamics \eqref{eq:controlled stochastic diffusion} using the Implicit-Euler integration scheme.  
\\
\indent We report the percentage of successful motions in Table \ref{table:robustness}. As shown, SNMPC executed all the motions successfully without violating the contact location constraints despite the disturbances, which satisfies the expected probability of constraint satisfaction ($99 \%$) thanks to the design of contact location constraints back-off design in \eqref{eq:chance-constraints backoffs}. On the contrary, NMPC violated the contact location constraints $48.3\%$ of all motions. Finally, HNMPC violated fewer constraints than NMPC, but still worse than SNMPC despite the robustness induced by shrinking the constraint set by hand. We highlight that although this heuristic works fairly for the trot case (success rate of $85.4\%$), using the same metric performed worse for a more agile bounding motion with a success rate of $67 \%$, which dictates that the user needs to keep tuning the controller blindly every time the OCP parameters changes to attain the desired empirical results. To quantify the safety margin induced by all controllers, we plot the mean and the $2\sigma$ distance between the end-effector positions and the center of the contact surface in Fig. \ref{fig:contact-location deviations norm} showing that SNMPC induced the best safety margin being the closest to the center of the contact surfaces.  
\\
\indent Finally, we plot the performance of the three controllers in Table \ref{table:performance} based on open-loop MPC, where we plug the predicted open-loop state instead of the measured state during re-planning. We highlight that we didn't use Monte-Carlo simulations in computing the performance due to the presence of a large number of failed trajectories in both NMPC and HNMPC cases. Although SNMPC sacrifices a bit the performance for safety, the performance of the three controllers is comparable. This is because the swing foot tracking of the controllers is affected by real-time iteration schemes (non-full-convergence of the OCPs), and the slack penalties on the constraints.  
\begin{figure}[!t]
    \begin{subfigure}[t]{\columnwidth}
        \begin{minipage}[b]{.24\columnwidth}
        \centering
        \includegraphics[trim=0 150 15 30, clip,scale=0.083]{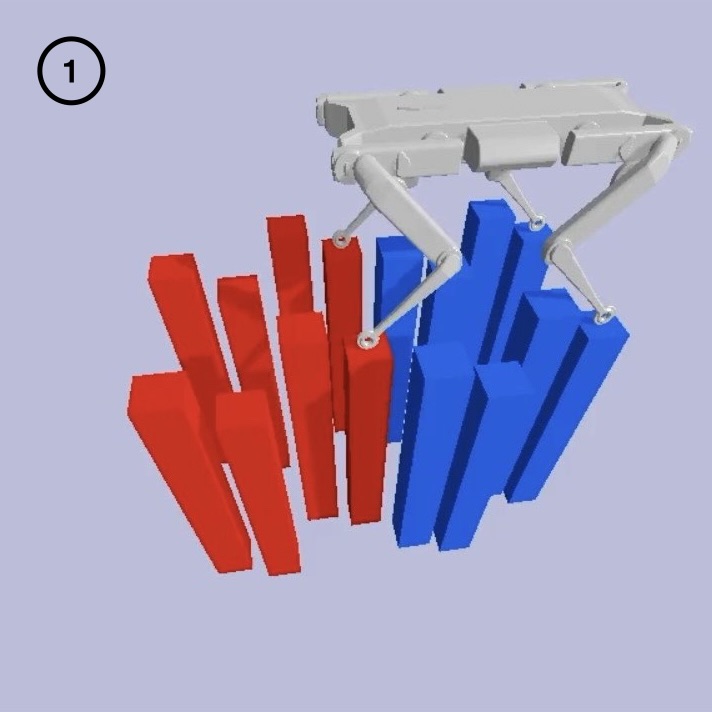}
        \end{minipage}
        \begin{minipage}[b]{.24\columnwidth}
        \includegraphics[trim=0 150 15 30 ,clip,scale=0.083]{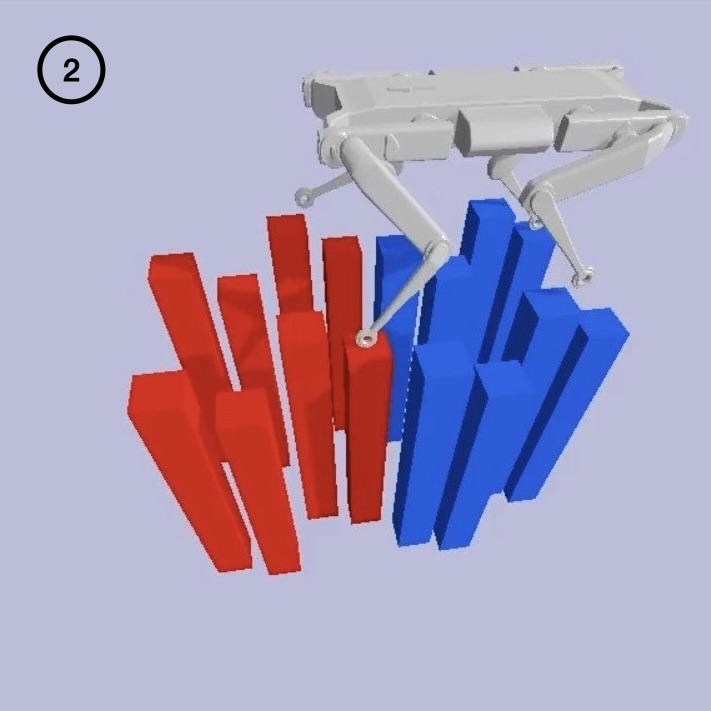}
        \end{minipage}
        \begin{minipage}[b]{.24\columnwidth}
        \centering
        \includegraphics[trim=0 150 15 30 ,clip,scale=0.083]{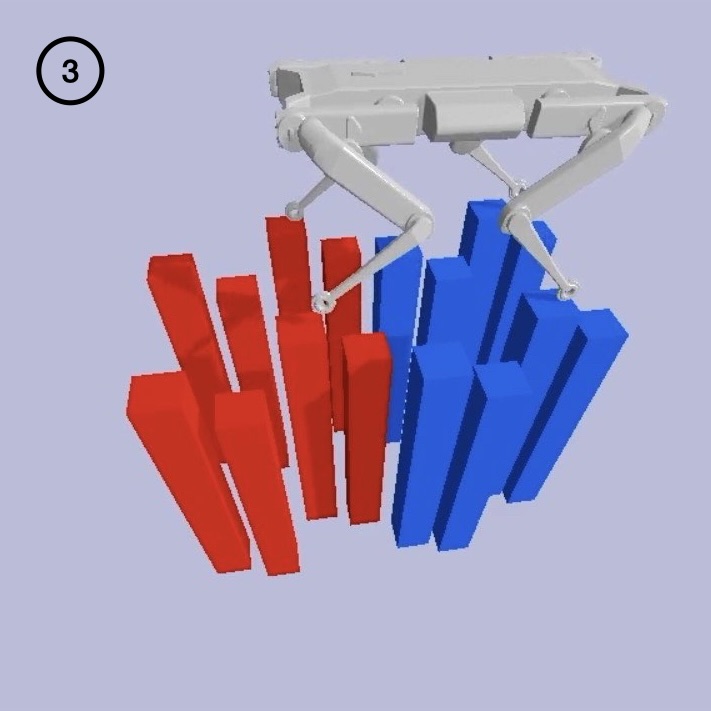}
        \end{minipage}
        \begin{minipage}[b]{.24\columnwidth}
        \centering
        \includegraphics[trim=0 150 15 30 ,clip,scale=0.083]{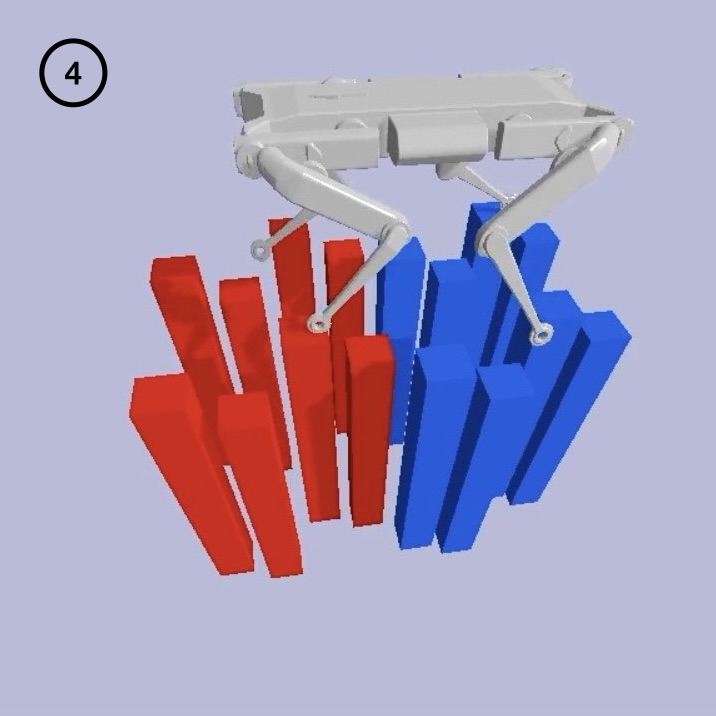}
        \end{minipage}
        \\
        \begin{minipage}[b]{.24\columnwidth}
        \centering
        \includegraphics[trim=0 150 15 30  ,clip,scale=0.083]{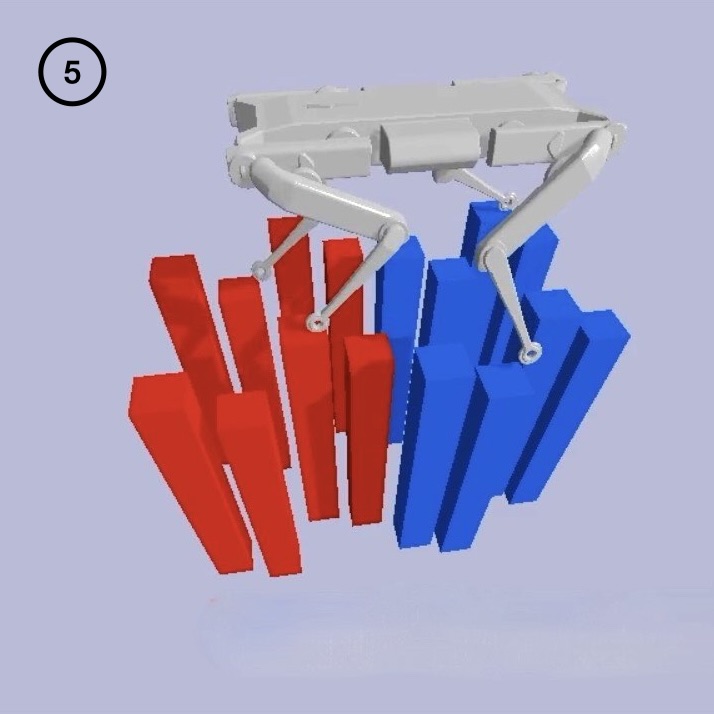}
        \end{minipage}
        \begin{minipage}[b]{.24\columnwidth}
        \includegraphics[trim=0 150 15 30  ,clip,scale=0.083]{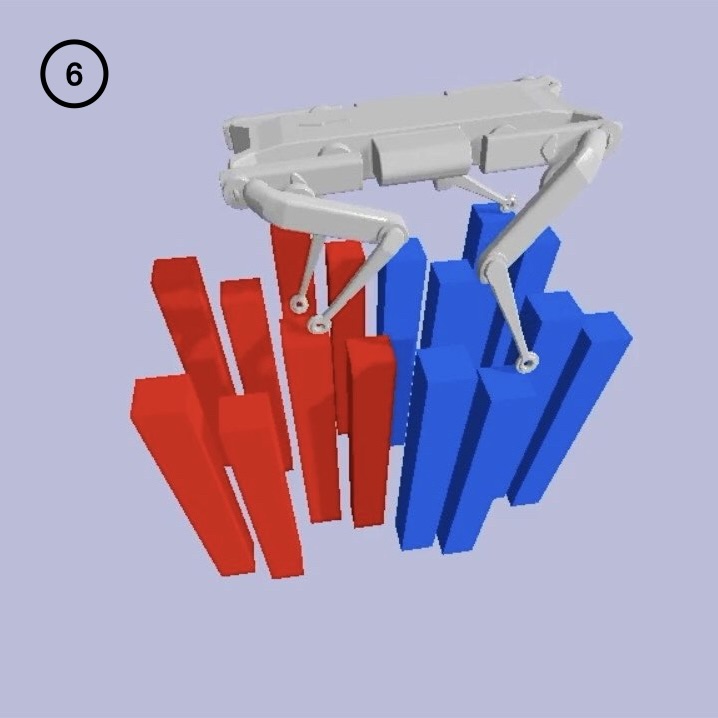}
        \end{minipage}
        \begin{minipage}[b]{.24\columnwidth}
        \centering
        \includegraphics[trim=0 150 15 30  ,clip,scale=0.083]{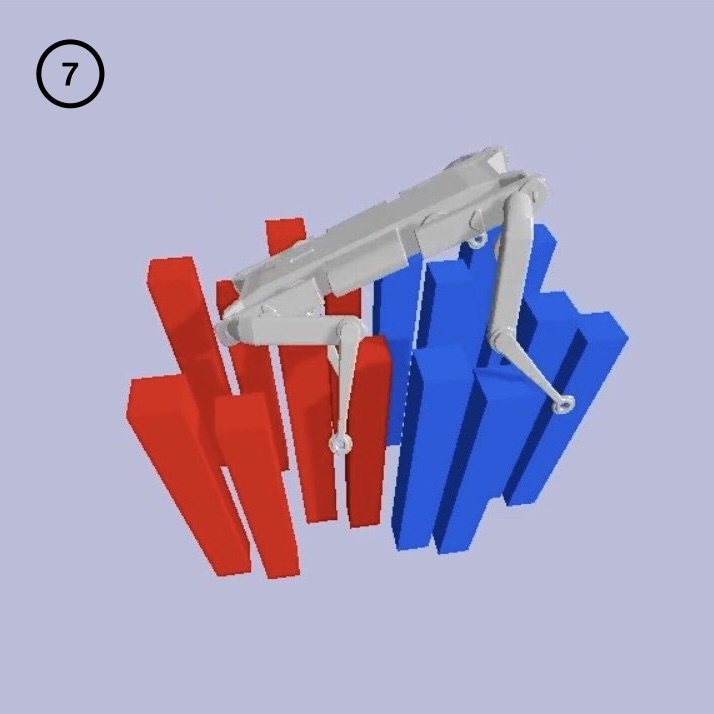}
        \end{minipage}
        \begin{minipage}[b]{.24\columnwidth}
        \centering
        \includegraphics[trim=0 150 15 30  ,clip,scale=0.083]{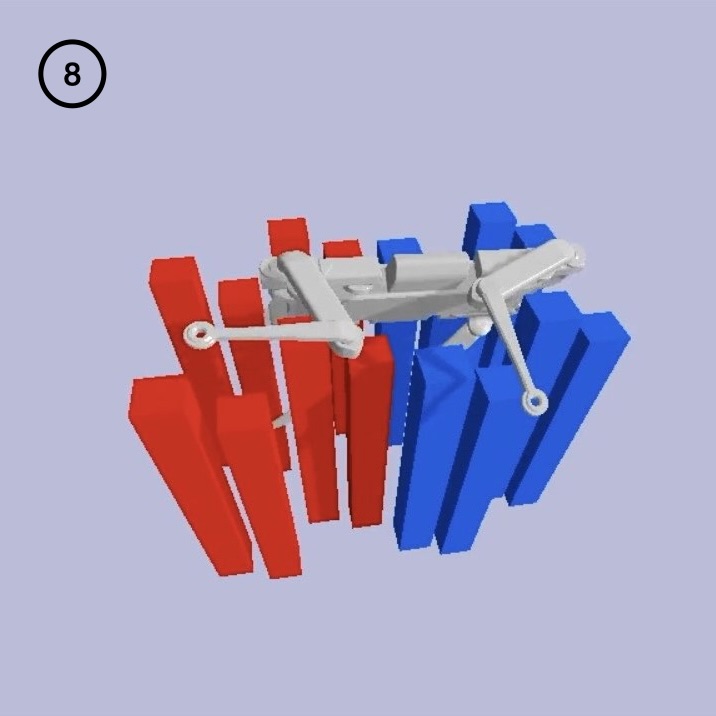}
        \end{minipage}
        \subcaption{Failed whole-body trot motion using NMPC.}
        \label{fig: trot NMPC snapshots}
        \end{subfigure}
    \begin{subfigure}[t]{\columnwidth}
        \begin{minipage}[b]{.24\columnwidth}
        \centering
        \includegraphics[trim=0 150 15 30  ,clip,scale=0.083]{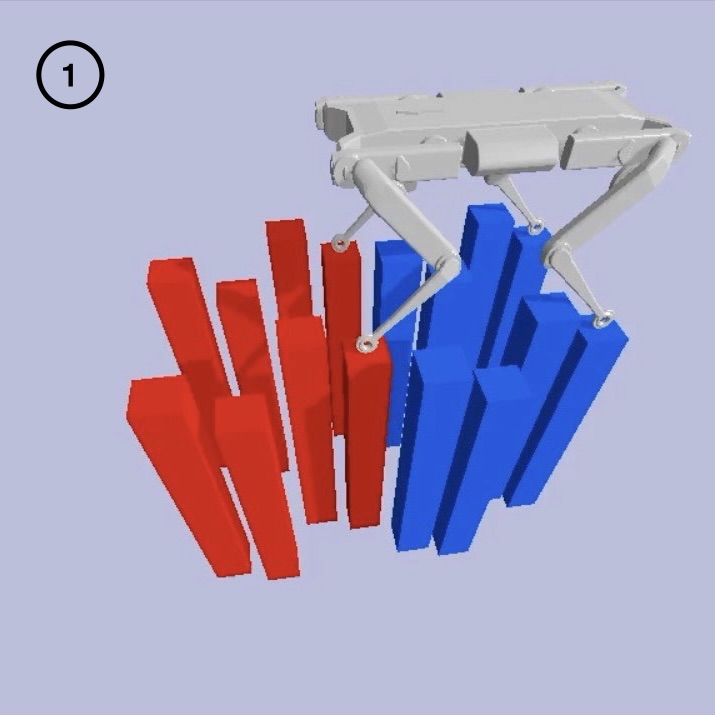}
        \end{minipage}
        \begin{minipage}[b]{.24\columnwidth}
        \includegraphics[trim=0 150 15 30  ,clip,scale=0.083]{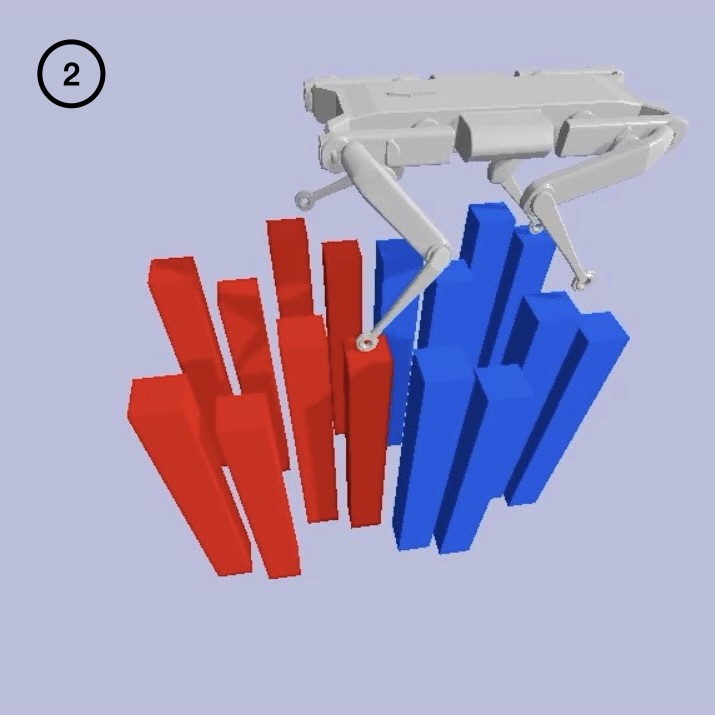}
        \end{minipage}
        \begin{minipage}[b]{.24\columnwidth}
        \centering
        \includegraphics[trim=0 150 15 30  ,clip,scale=0.083]{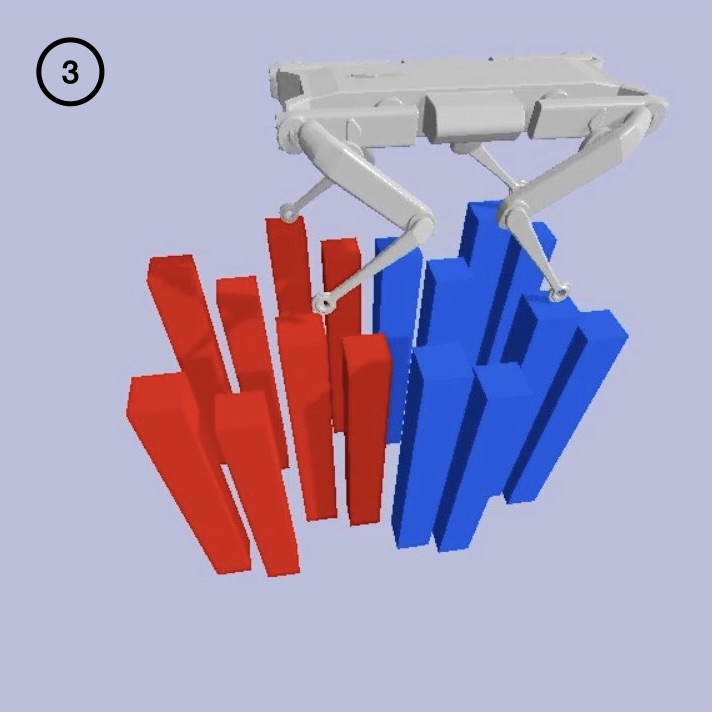}
        \end{minipage}
        \begin{minipage}[b]{.24\columnwidth}
        \centering
        \includegraphics[trim=0 150 15 30  ,clip,scale=0.083]{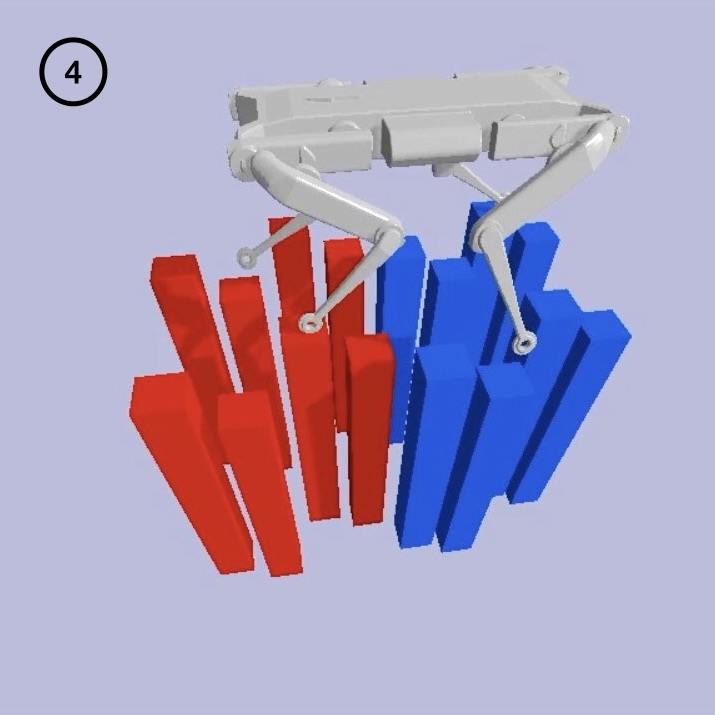}     
        \end{minipage}
        \\
        \begin{minipage}[b]{.24\columnwidth}
        \centering
         \includegraphics[trim=0 150 15 30  ,clip,scale=0.083]{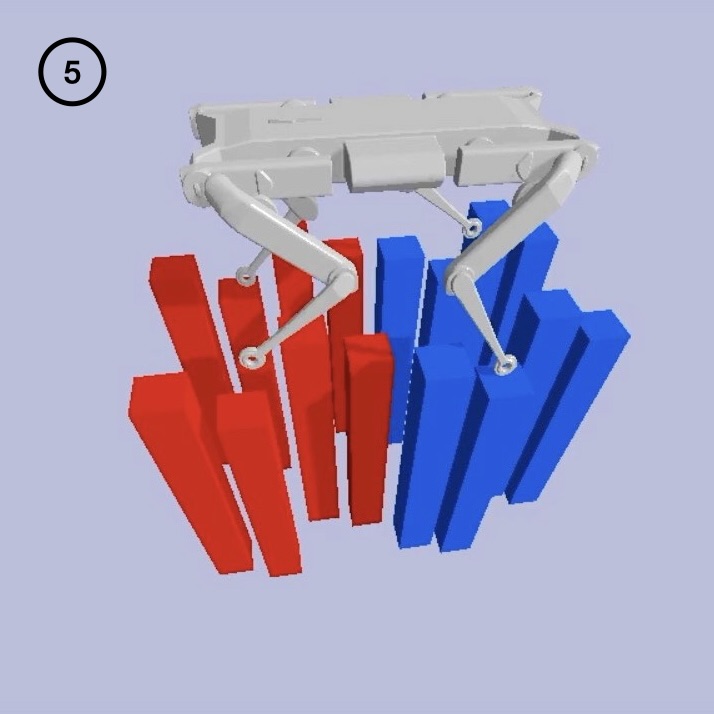}
        \end{minipage}
        \begin{minipage}[b]{.24\columnwidth}
         \includegraphics[trim=0 150 15 30  ,clip,scale=0.083]{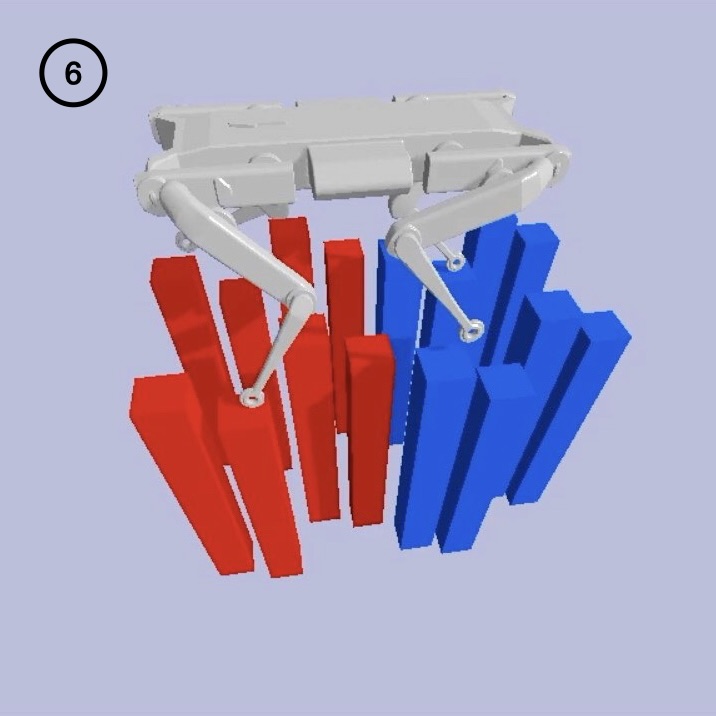}
        \end{minipage}
        \begin{minipage}[b]{.24\columnwidth}
        \centering
        \includegraphics[trim=0 150 15 30  ,clip,scale=0.083]{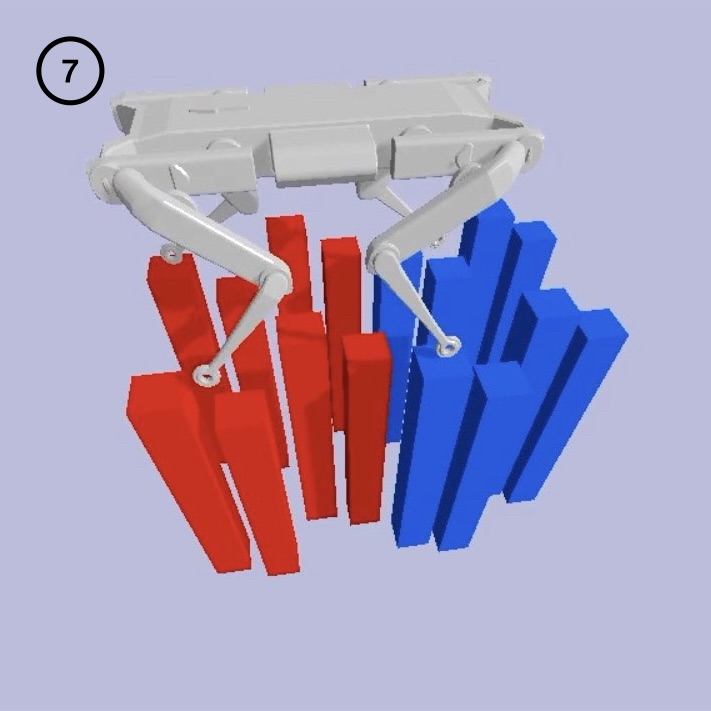}
        \end{minipage}
        \begin{minipage}[b]{.24\columnwidth}
        \centering
         \includegraphics[trim=0 150 15 30  ,clip,scale=0.083]{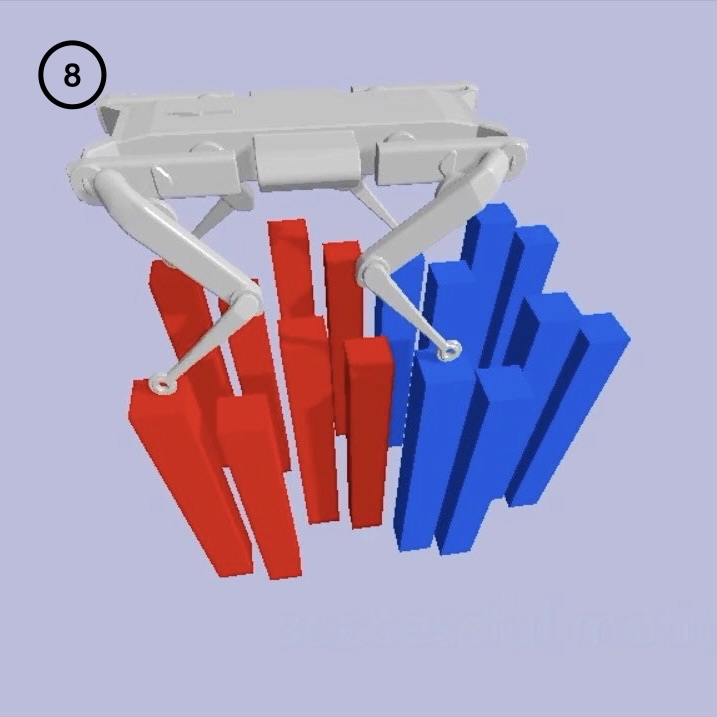}
        \end{minipage}
        \subcaption{Successful whole-body trot motion using SNMPC.}
        \label{fig: trot SMPC snapshots}
    \end{subfigure}
    \caption{Snapshots of a trot motion on non-coplanar stones.}
    \label{fig: trot motion}
\end{figure}
%
\subsection{Whole-body Simulation}
\begin{figure}[!t]
    \begin{subfigure}[t]{\columnwidth}
        \begin{minipage}[b]{.24\columnwidth}
        \centering
        \includegraphics[trim=0 150 15 30  ,clip,scale=0.083]{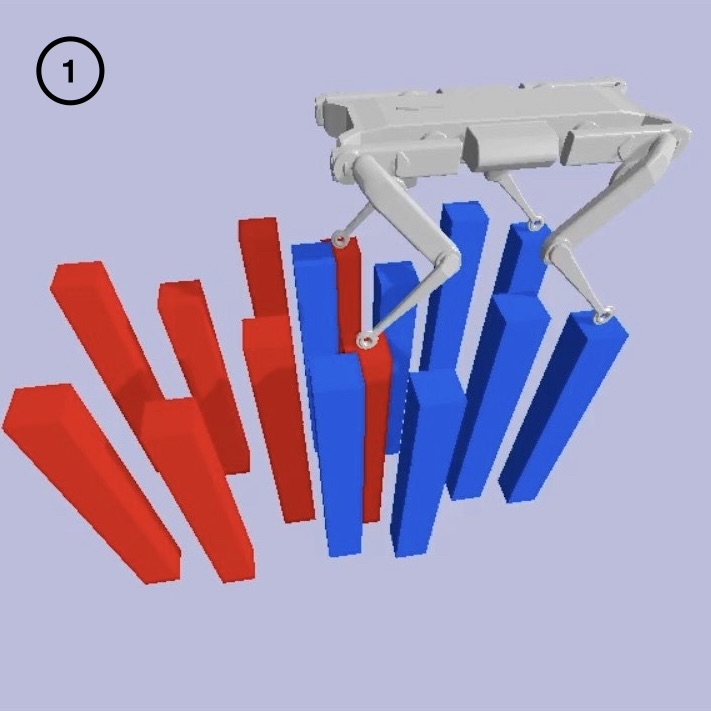}
        \end{minipage}
        \begin{minipage}[b]{.24\columnwidth}
        \includegraphics[trim=0 150 15 30  ,clip,scale=0.083]{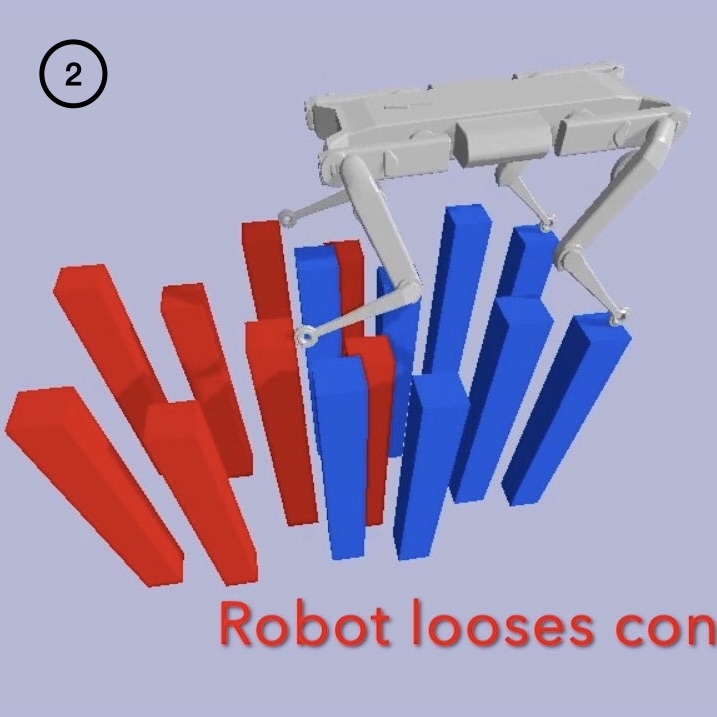}
        \end{minipage}
        \begin{minipage}[b]{.24\columnwidth}
        \centering
       \includegraphics[trim=0 150 15 30  ,clip,scale=0.083]{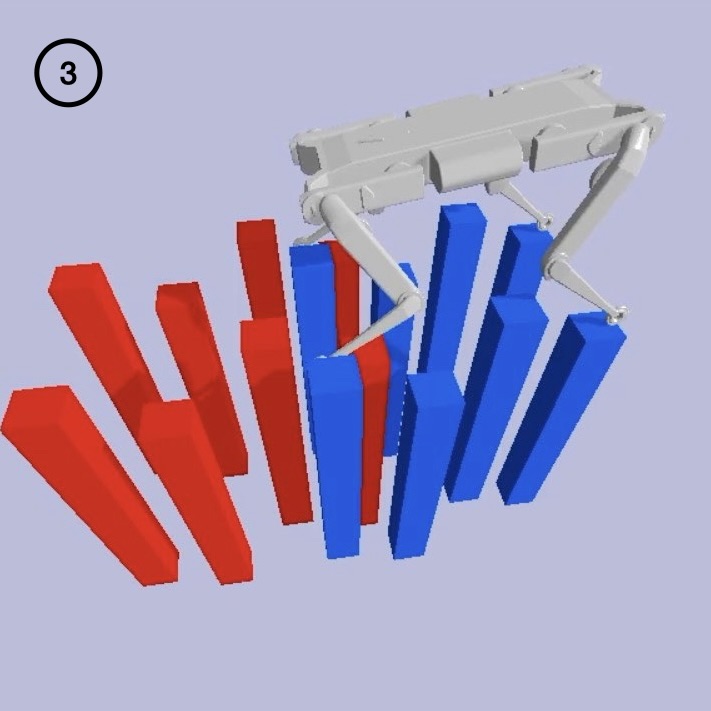}
        \end{minipage}
        \begin{minipage}[b]{.24\columnwidth}
        \centering
        \includegraphics[trim=0 150 15 30  ,clip,scale=0.083]{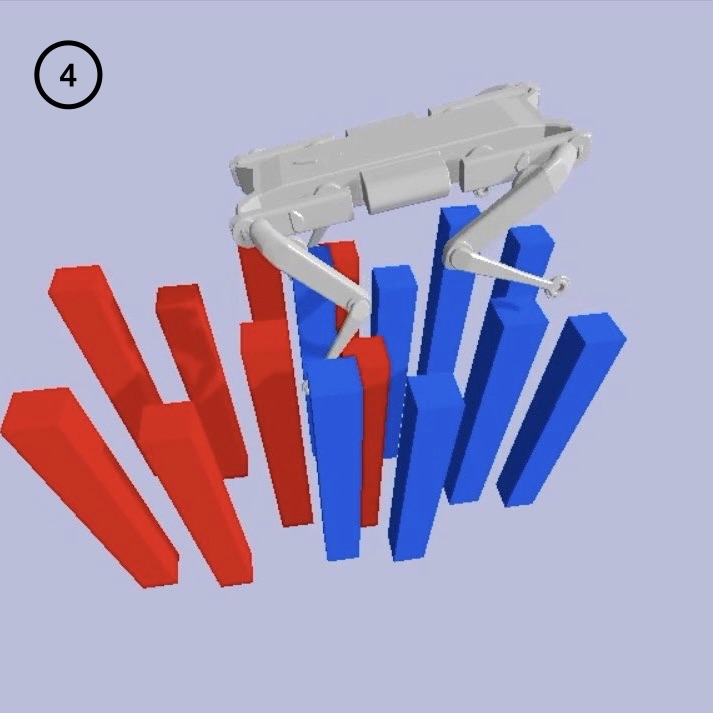}
        \end{minipage}
        \\
        \begin{minipage}[b]{.24\columnwidth}
        \centering
       \includegraphics[trim=0 150 15 30  ,clip,scale=0.083]{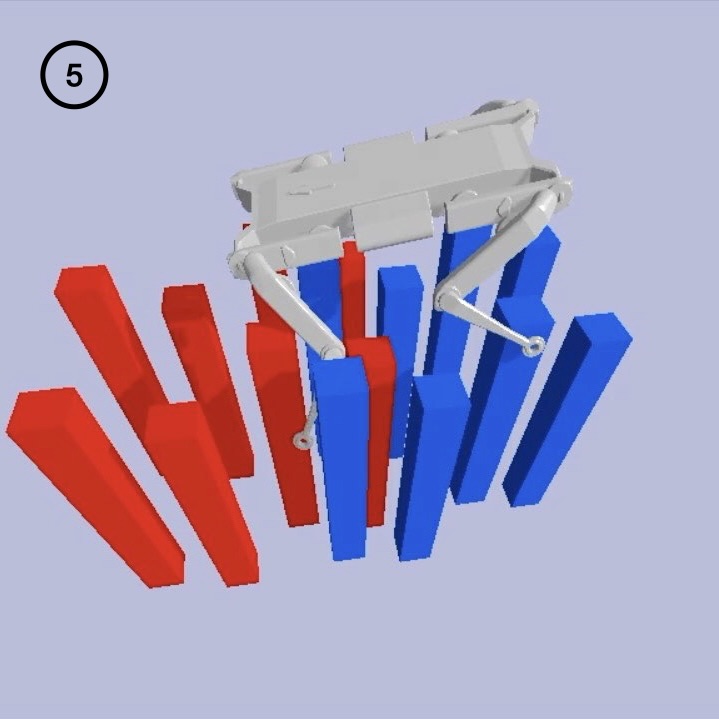}
        \end{minipage}
        \begin{minipage}[b]{.24\columnwidth}
        \includegraphics[trim=0 150 15 30  ,clip,scale=0.083]{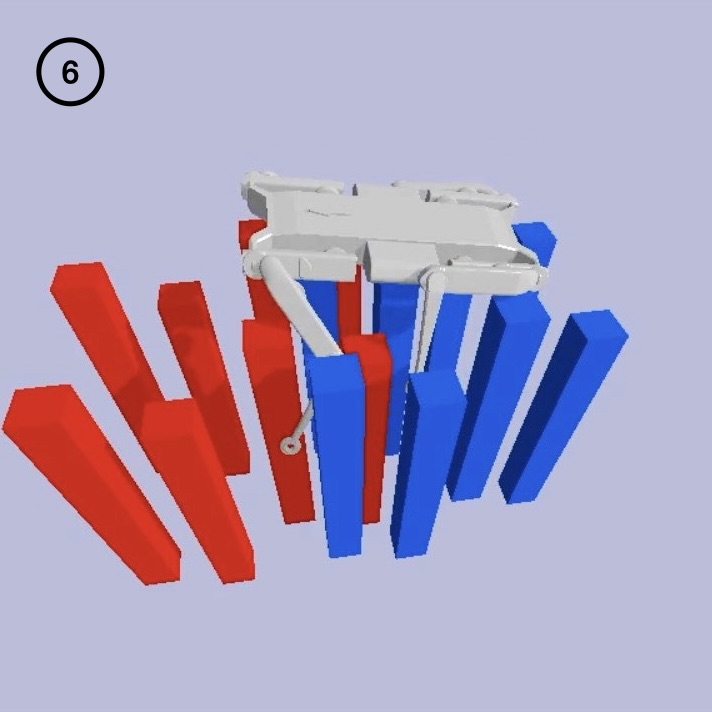}
        \end{minipage}
        \begin{minipage}[b]{.24\columnwidth}
        \centering
        \includegraphics[trim=0 150 15 30  ,clip,scale=0.083]{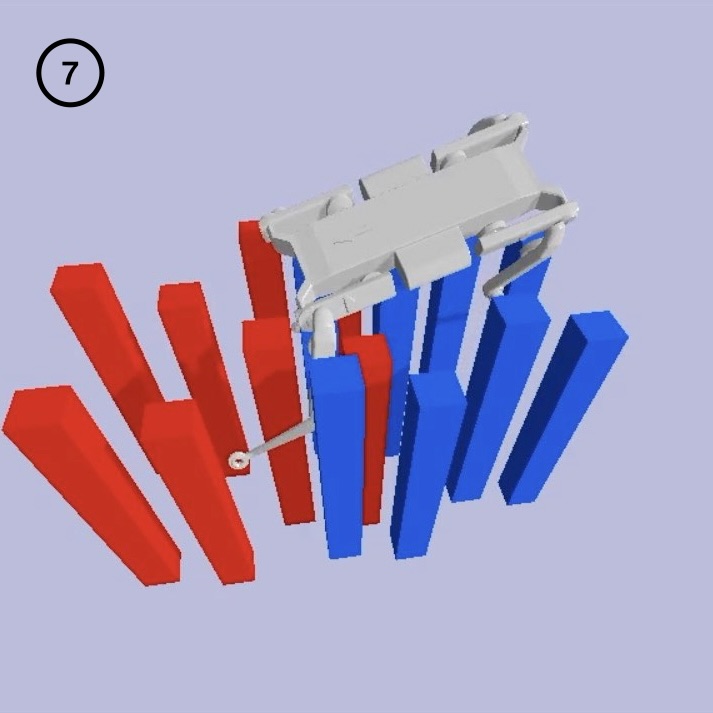}
        \end{minipage}
        \begin{minipage}[b]{.24\columnwidth}
        \centering
       \includegraphics[trim=0 150 15 30  ,clip,scale=0.083]{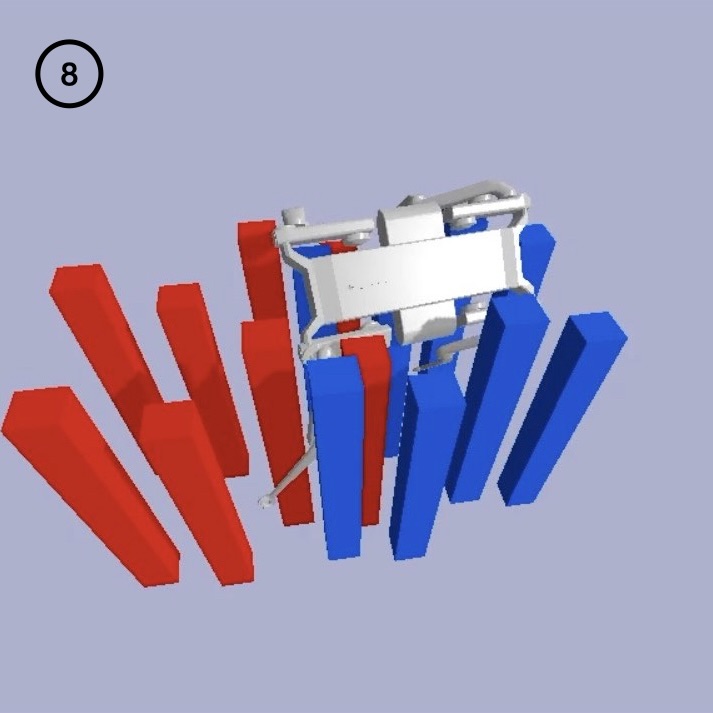}
        \end{minipage}
        \subcaption{Failed whole-body bounding motion using NMPC.}
        \label{fig: bound NMPC snapshots}
    \end{subfigure}
    \begin{subfigure}[t]{\columnwidth}
        \begin{minipage}[b]{.24\columnwidth}
        \centering
        \includegraphics[trim=0 150 15 30  ,clip,scale=0.083]{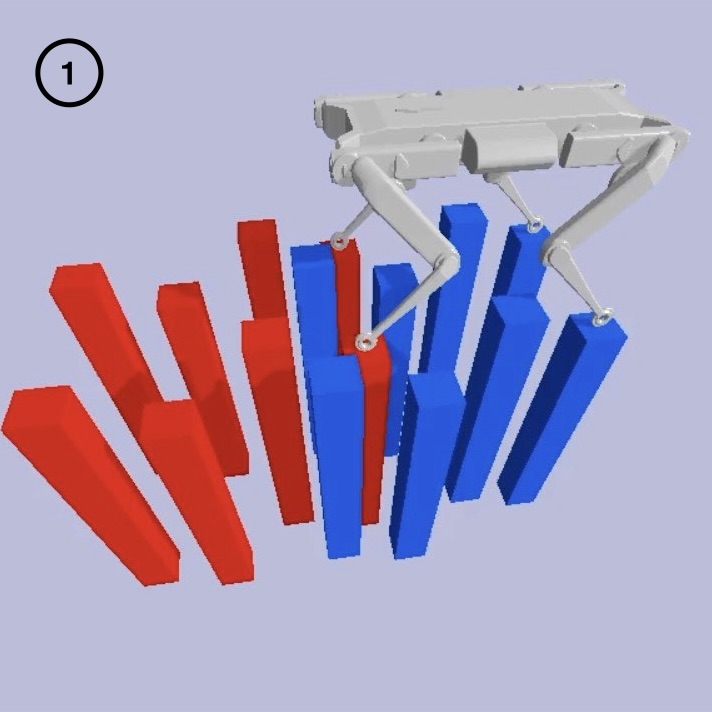}
        \end{minipage}
        \begin{minipage}[b]{.24\columnwidth}
         \includegraphics[trim=0 150 15 30  ,clip,scale=0.083]{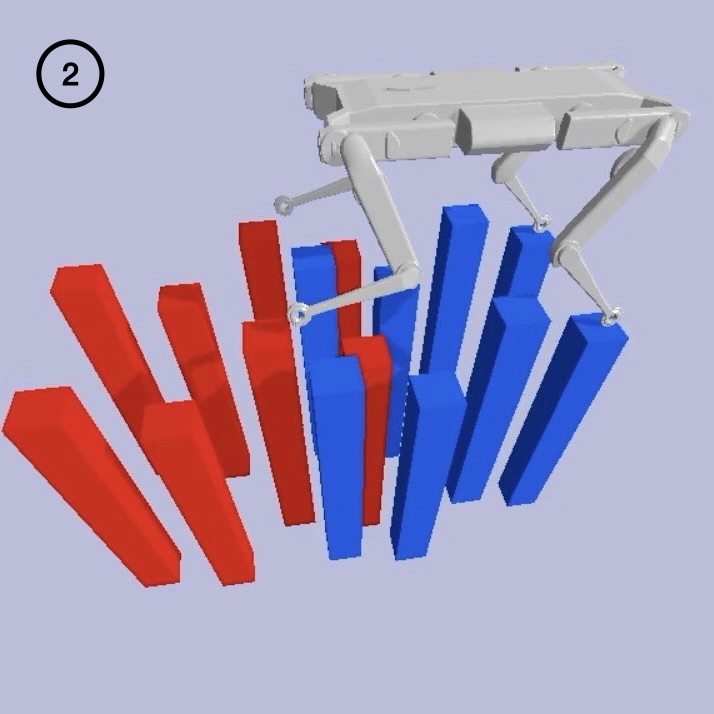}
        \end{minipage}
        \begin{minipage}[b]{.24\columnwidth}
        \centering
       \includegraphics[trim=0 150 15 30  ,clip,scale=0.083]{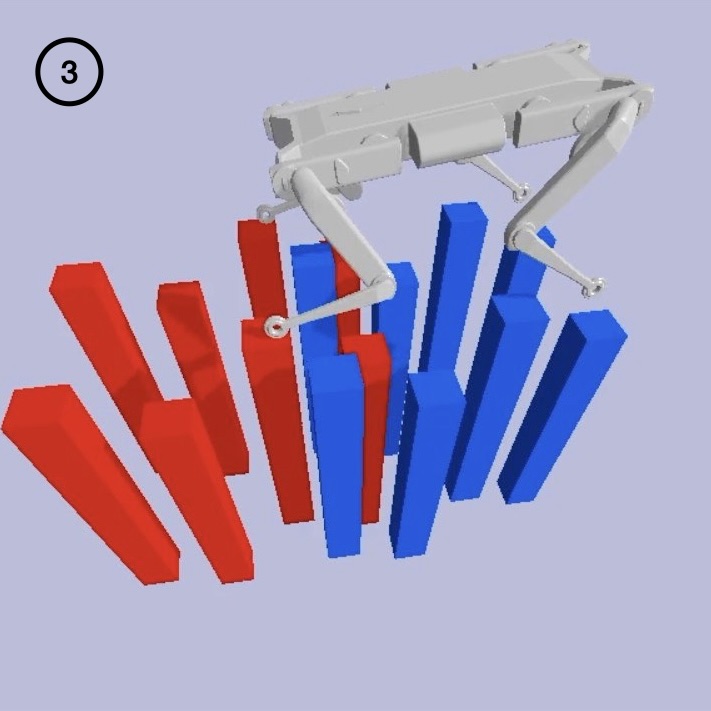}
        \end{minipage}
        \begin{minipage}[b]{.24\columnwidth}
        \centering
        \includegraphics[trim=0 150 15 30  ,clip,scale=0.083]{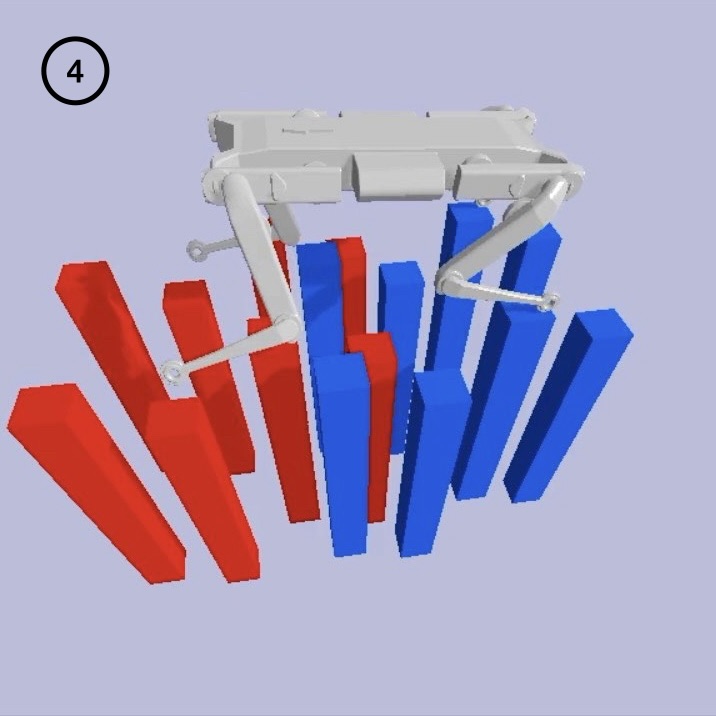}
        \end{minipage}
        \\
        \begin{minipage}[b]{.24\columnwidth}
        \centering
       \includegraphics[trim=0 150 15 30  ,clip,scale=0.083]{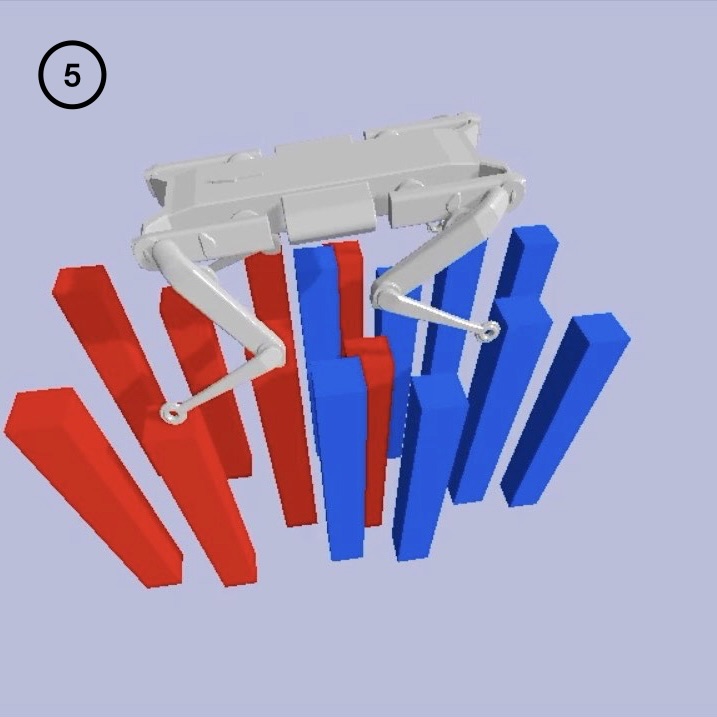}
        \end{minipage}
        \begin{minipage}[b]{.24\columnwidth}
       \includegraphics[trim=0 150 15 30  ,clip,scale=0.083]{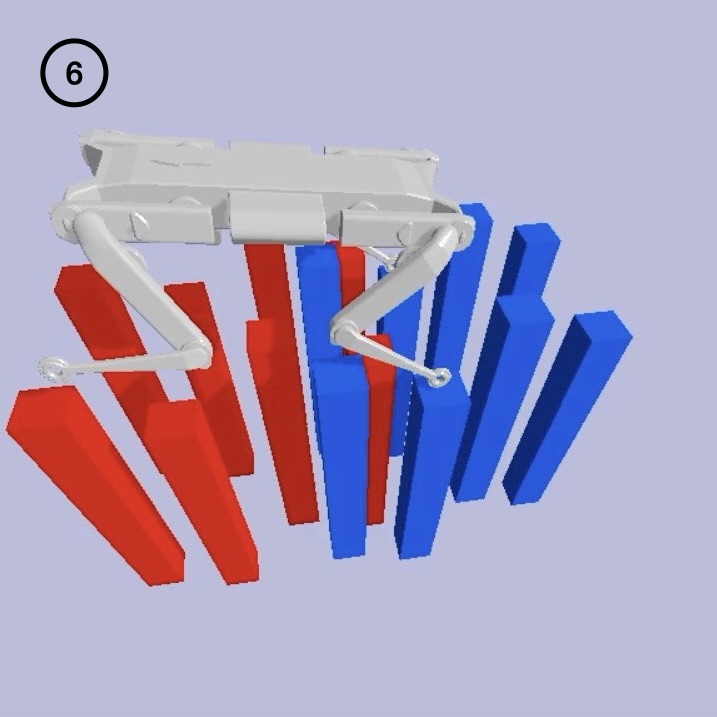}
        \end{minipage}
        \begin{minipage}[b]{.24\columnwidth}
        \centering
       \includegraphics[trim=0 150 15 30  ,clip,scale=0.083]{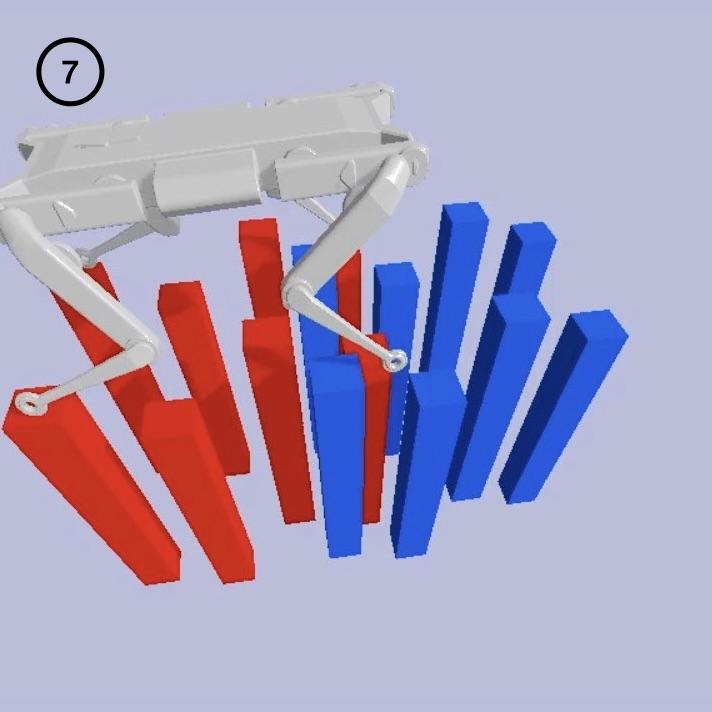}
        \end{minipage}
        \begin{minipage}[b]{.24\columnwidth}
        \centering
       \includegraphics[trim=0 150 15 30  ,clip,scale=0.083]{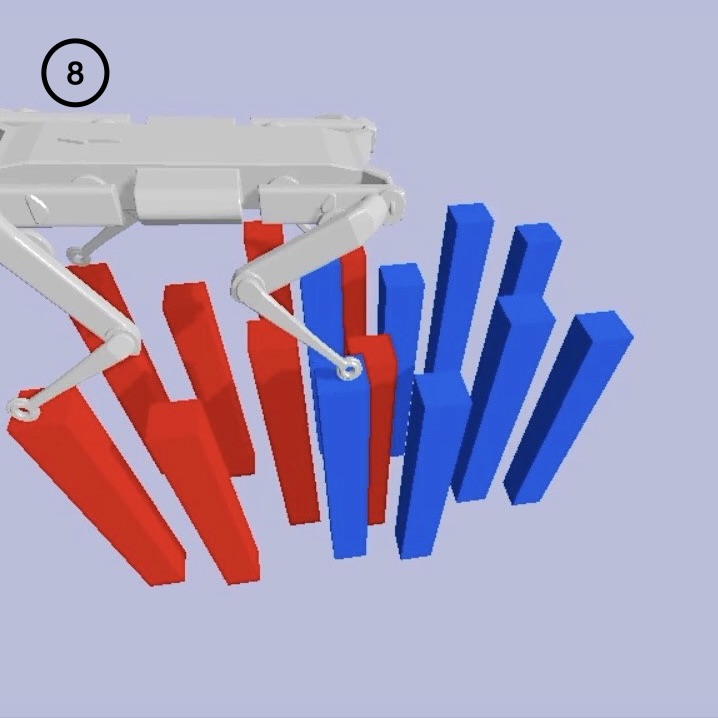}
        \end{minipage}
        \subcaption{Successful whole-body bounding motion using SNMPC.}
        \label{fig: bound SMPC snapshots}
    \end{subfigure}
    \caption{\hspace{-0.01cm}Snapshots of a bound motion on non-coplanar stones.}
    \label{fig: bound motion}
\end{figure}

In this subsection, we test the effect of model mismatch between the kino-dynamic model and the whole-body model (i.e. the dynamic effects of the legs). Although in practice the legs are assumed to be massless for quadruped robots, their effect cannot be neglected for agile motions. Moreover, impulsive dynamics during impacts are also ignored since they are usually hard to model. Finally, since we are running real-time iterations, neither solver achieves full convergence in one Newton/Newton-type step. As a consequence, the previous effects can hinder the satisfaction of the contact location constraints. To test those effects, we report whole-body simulations of the quadruped robot Solo ~\cite{grimminger2020open} in the Pybullet simulation environment \cite{coumans2016pybullet} for dynamic trot and bound motions shown in Fig. \ref{fig: trot motion} and Fig. \ref{fig: bound motion} respectively. The whole-body simulation runs with a discretization time of $\Delta_{sim_k} = 1$ ms, where the feedforward MPC trajectories are linearly interpolated. We apply the following state feedback control law to both controllers:
\begin{IEEEeqnarray}{lcr}
\label{eq:wholebody control law}
    \boldsymbol{\tau}_k = \boldsymbol{\tau}^\ast_k + \boldsymbol{K}_{p}(\boldsymbol{q}^\ast_{j_k} - \boldsymbol{q}_{j_k}) + \boldsymbol{K}_{d}(\dot{\boldsymbol{q}}_{j_k} - \dot{\boldsymbol{q}}_{j_k}), \IEEEyesnumber\IEEEyessubnumber 
    \\
    \boldsymbol{\tau}^\ast_k  \triangleq \mathrm{RNEA}(\tilde{\boldsymbol{{q}}}^\ast_k, \dot{\boldsymbol{q}}^\ast_k, \ddot{\boldsymbol{q}}^\ast_k) - \sum^{n_c}_{i=0} \boldsymbol{J}^\top_i(\tilde{\boldsymbol{q}}^\ast_k)\boldsymbol{\lambda}^\ast_{i}.\IEEEyessubnumber
\end{IEEEeqnarray}
The feedforward torques are computed using  the Recursive Newton-Euler Algorithm ($\mathrm{RNEA}$) ~\cite{featherstone2014rigid}. The joint position and velocity feedback gains are set to $\boldsymbol{K}_p = 2.\mathbb{I}_{n_j \times n_j}$ and $\boldsymbol{K}_d = 0.15.\mathbb{I}_{n_j \times n_j}$ respectively. The superscript $^\ast$ represents the optimized quantities coming from MPC. For the trot motion, NMPC and HNMPC failed to complete the motion by breaking contact in the second step as shown in Fig. \ref{fig: trot NMPC snapshots}. On the contrary, SNMPC manages to complete the motion successfully until the end (see Fig. \ref{fig: trot SMPC snapshots}). We tested the effect of leg inertia for an agile bounding motion, where NMPC and HNMPC failed again during the second bounding step (see Fig.~\ref{fig: bound NMPC snapshots}), while SNMPC successfully completed the motion as shown in Fig.~\ref{fig: bound SMPC snapshots} (check the submission video). 
%

\section{Discussion and Conclusions}
In this work, we tackled the problem of kino-dynamic stochastic trajectory optimization subject to additive uncertainties on the dynamics and contact location chance constraints. We designed contact location safety constraints by computing upper bounds (back-offs) that take into account the linearized propagated uncertainties along the planning horizon assuming a Gaussian distribution of those uncertainties. The final solution is an approximation of the original SNMPC problem solved using a real-time iteration scheme. We compared the robustness of SNMPC against NMPC by running 500 Monte-Carlo kino-dynamic simulations for agile trotting and bounding motions for the quadruped robot Solo on a challenging non-coplanar environment with small stepping stones as well as whole-body simulations. SNMPC completed all the motions successfully without violating the contact locations constraints, while NMPC violated them $48.3\%$ of the time. Moreover, we ran whole-body simulations in Pybullet to study the effects of mismatch between the kino-dynamic and whole-body model; SNMPC was able to complete both motions successfully, while NMPC failed in both cases showing the benefit of SNMPC over deterministic planning in safety-critical scenarios. 
\\
\indent Finally, we compared the robustness of SNMPC against HNMPC. Since the robustness of SNMPC is induced by designing proper back-offs, then one might think why not design such safety margins heuristically by shrinking the constraint set by hand? We argued that although this approach might work in practice for some cases, it does not provide an automatic procedure for designing such safety margins leaving this to a process of trial and error. For instance, what should be the proper safety margin for different agile motion plans without degrading performance? As shown in our empirical results, using the same heuristic safety margin for both trotting and bounding motions yielded different safety rates of successful motions. Moreover, this heuristic-based approach does not relate the magnitude of back-off design with the uncertainty statistics that might be available from previously collected data about the system in simulation or on the real robot. On the contrary, SNMPC methodologically addresses those issues by computing such bounds automatically, which vary at each point in time based on expected uncertainty propagation along the horizon, the time-varying closed-loop feedback gain,  and the desired probability of satisfying such constraints \eqref{eq:chance-constraints backoffs}.
\\
\indent Regarding the tuning of SNMPC, we did not re-tune different motions as it comes automatically by design due to consideration of the model uncertainties. Most of the tuning effort was allocated to design the baseline motions with NMPC using the slack penalties for constraint satisfaction along with the cost function weights, which is a classic effort for tuning constrained nonlinear trajectory optimizers.
\\
\indent One limitation of the current work is that it does not take into account contact mode uncertainties, which are of combinatorial nature. We would like to explore tractable SNMPC formulations that take into account contact time uncertainties induced by uncertainties in the discrete contact modes, which is beneficial for sequential manipulation and locomotion tasks. Moreover, we intend to test the current SNMPC scheme on real robot experiments in future work.  

    \addtolength{\textheight}{0cm}     
    \bibliographystyle{IEEEtran}
    \bibliography{IEEEabrv,Biblio}
\end{document}